\documentclass[10pt,twocolumn,letterpaper]{article}

\usepackage[pagenumbers]{cvpr} 

\usepackage{currfile} 

\usepackage{caption} 

\usepackage{makecell}
\usepackage{placeins}

\usepackage{algorithm}
\usepackage{algpseudocode}

\usepackage{booktabs}
\usepackage{array}
\usepackage{adjustbox} 

\usepackage{enumitem}
\setlist[itemize,enumerate]{topsep=3pt, itemsep=2pt, parsep=0pt, leftmargin=1.5em}

\usepackage[most]{tcolorbox}
\newtcolorbox{promptbox}[2][]{
  enhanced,
  breakable,
  colback=black!2,
  colframe=cvprblue!60!black,
  colbacktitle=cvprblue!85!black,
  coltitle=white,
  fonttitle=\bfseries\footnotesize,
  title={#2},
  boxrule=0.5pt,
  arc=2pt,
  left=6pt, right=6pt, top=5pt, bottom=5pt,
  toptitle=3pt, bottomtitle=3pt,
  before skip=6pt, after skip=6pt,
  fontupper=\ttfamily\scriptsize\raggedright,
  #1
}

\newcolumntype{L}[1]{>{\raggedright\arraybackslash}p{#1}}

\usepackage{float}

\newcommand{\MATCHAppendixFloats}{%
  \renewcommand{\textfraction}{0.10}
  \renewcommand{\topfraction}{0.90}
  \renewcommand{\bottomfraction}{0.85}
  \renewcommand{\dbltopfraction}{0.90}
  \renewcommand{\floatpagefraction}{0.75}
  \renewcommand{\dblfloatpagefraction}{0.75}
  \setcounter{dbltopnumber}{3}
  \setlength{\dbltextfloatsep}{12pt plus 2pt minus 2pt}
  \setlength{\dblfloatsep}{8pt plus 2pt minus 2pt}
  \setlength{\textfloatsep}{12pt plus 2pt minus 2pt}
  \setlength{\floatsep}{8pt plus 2pt minus 2pt}
}
\ifdefined\MATCHSupplementary\MATCHAppendixFloats\fi

\makeatletter
\@ifpackageloaded{lineno}{\ifdefined\IfFormatAtLeastTF\IfFormatAtLeastTF{2025-06-01}{%
  \@ifpackagelater{lineno}{2026/03/08}{}{\@ifpackagelater{lineno}{2026/02/16}{%
    \RemoveFromHook{build/column/before}[lineno]%
    \AddToHook{build/column/after}{%
      \setbox\@outputbox\vbox to\ht\@outputbox{%
        \boxmaxdepth\@maxdepth
        \protected@write\@auxout{}{%
          \string\def\string\@LN@column{\if@firstcolumn1\else2\fi}}%
        \unvbox\@outputbox}}%
  }{}}}{}\fi}{}
\makeatother

\definecolor{cvprblue}{rgb}{0.21,0.49,0.74}
\usepackage[pagebackref,breaklinks,colorlinks,allcolors=cvprblue]{hyperref}

\def\confName{CVPR}
\def\confYear{2027}

\title{RoboMATCH: Monitoring-Augmented Task and Control Hierarchy for Adaptive Multi-Timescale Coordination in Generalist Robot Manipulation}

\author{
Zihan Yang$^{1,2,*}$ \quad Zhangzheng Tu$^{1,2,*}$ \quad Weiye Hu$^{3}$ \quad Baijun Chen$^{1,2}$ \\
Zanxin Chen$^{1,2}$ \quad Xiangwei Jiang$^{1}$ \quad Xiaokang Yang$^{1}$ \quad Yao Mu$^{1,2,\dagger}$ \\[4pt]
{\small $^{1}$Shanghai Jiao Tong University \quad $^{2}$Shanghai AI Laboratory} \\
{\small $^{3}$Xiamen University Malaysia} \\[2pt]
{\small $^{*}$Equal contribution. \quad $^{\dagger}$Corresponding author: \texttt{muyao@sjtu.edu.cn}.}
}

\makeatletter
\let\MATCHOriginalTitle\@maketitle
\def\@maketitle{\MATCHOriginalTitle\vspace{6pt}\begingroup
    \centering
    \includegraphics[width=0.98\linewidth]{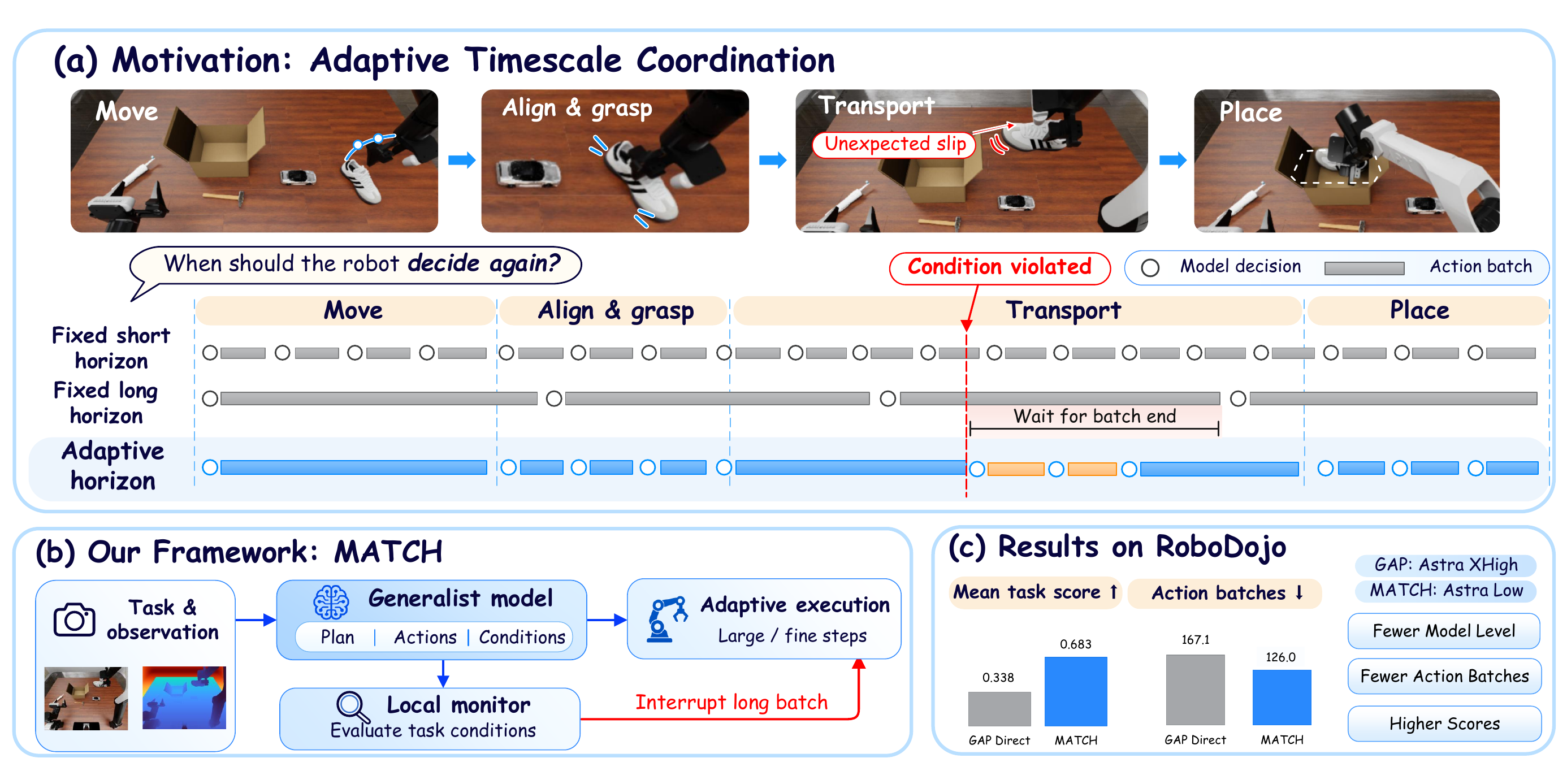}
    \vspace{-4pt} 
    \captionsetup{hypcap=false}
    \captionof{figure}{
        \textbf{Motivation and overview of MATCH.}
        (a) \textbf{Motivation.} Fixed short horizons incur redundant queries, whereas fixed long horizons delay reactions to unexpected events (e.g., object slip). MATCH adapts decision timing to physical states—sustaining \textit{large-step} batches (blue) in stable stages while immediately switching to \textit{fine-step} recovery (orange) when conditions are violated.
        (b) \textbf{Framework.} A single generalist model jointly generates subtask plans, adaptive-horizon actions, and executable conditions, while a control-rate local monitor evaluates task states to trigger mid-batch interruption and feedback-driven replanning.
        (c) \textbf{Results.} On RoboDojo, MATCH outperforms GAP Direct in both task performance and decision efficiency, even with lower reasoning effort.
    }
    \captionsetup{hypcap=true}
    \label{fig:teaser}
\par\endgroup\vspace{6pt}}
\makeatother
\begin{document}
\maketitle

\begin{abstract}
While general-purpose large models increasingly demonstrate the ability to directly generate robot actions, their high inference latency creates a fundamental timescale dilemma in continuous physical execution: short execution horizons incur prohibitive decision overhead, whereas fixed long horizons suffer from open-loop blindness during fine-grained contact or unexpected anomalies. To resolve this conflict, we propose \textbf{MATCH} (Monitoring-Augmented Task and Control Hierarchy), a closed-loop multi-timescale framework for direct end-effector control that dynamically adapts high-level decision frequency to low-level physical states. Within MATCH, a single generalist model jointly synthesizes reusable subtask plans, multi-granularity action batches, and executable task-conditioned monitoring criteria. During execution, a lightweight local monitor evaluates physical conditions at control-step resolution—sustaining long-horizon execution in stable free-space phases while triggering immediate mid-batch interruption and feedback-driven replanning upon anomalies or fine-alignment needs. Evaluated across six RoboDojo manipulation tasks, MATCH boosts the average task score from 0.338 to 0.683 while reducing action decision batches by 24.6\%, outperforming direct model control even with lower reasoning effort. Ablations confirm that these gains stem from the tight closed-loop coupling of planning, adaptive horizons, and runtime monitoring rather than merely reducing query frequency.
\end{abstract}
\section{Introduction}
\label{sec:intro}

In recent years, general-purpose large models have made significant progress in language understanding, visual perception, spatial reasoning, and tool use~\cite{OpenAI2023GPT4TR,Bai2023QwenTR}. Their application to robotics has also gradually extended from high-level task planning to low-level action generation~\cite{Driess2023PaLMEAE,Brohan2023RT2VM,Zhang2026AnUR}. From decomposing complex tasks using language models~\cite{Ichter2022DoAI}, to understanding scenes and developing operation plans with vision-language models~\cite{Hu2023LookBY,Driess2023PaLMEAE}, and further to directly generating end-effector targets or robot actions~\cite{Liang2022CodeAP,Brohan2023RT2VM,Zhang2026AnUR}, general-purpose large models are increasingly demonstrating the potential to control robots directly.

These advances also introduce a new challenge: \textbf{robots must not only generate correct actions, but also make decisions at appropriate timescales.} In language interaction, users can wait for a model to complete its reasoning; in robotic tasks, however, the system must continuously respond during motion, contact, and environmental changes. We observe that, when GPT-6 Astra is directly used as a robot policy, model calls account for more than 90\% of the total task time, severely limiting execution efficiency. To alleviate this conflict, recent hierarchical robotic systems organize semantic reasoning, action generation, and low-level control at different timescales~\cite{Shi2025HiRO}, allowing costly high-level decisions to be reused across multiple control steps.

However, \textbf{dividing a system into modules operating at different timescales does not necessarily enable efficient coordination among these levels.} Fixed action execution horizons are difficult to adapt to the needs of different task stages. In relatively stable stages, such as free-space motion, frequent calls to large models cause unnecessary inference overhead. During grasp alignment, contact manipulation, or unexpected events, however, robots require more timely feedback and policy adjustment. Furthermore, methods that primarily rely on subtask completion signals to trigger replanning~\cite{Chen2026CanAS} may overlook anomalies and risks arising during execution. Thus, the key is not only how to assign operating frequencies to different levels, but also \textbf{how to dynamically coordinate high-level decision frequency and low-level execution behavior according to task progress and actual execution states.} A robot should fully reuse existing decisions and execute longer action batches during stable stages, while obtaining timely feedback and adjusting subsequent actions when fine-grained correction is required or execution anomalies occur, thereby balancing execution efficiency and responsiveness.

Based on this observation, we propose \textbf{MATCH} (Monitoring-Augmented Task and Control Hierarchy), a closed-loop multi-timescale coordination framework for direct generalist end-effector control. Rather than orchestrating separate planners or external failure detectors, MATCH uses a single generalist model to jointly synthesize and selectively update reusable subtask plans, operation specifications, multi-granularity action batches, and executable monitoring conditions. During execution, the controller dynamically alternates between \textit{large-step} end-effector pose waypoints—such as rapidly traversing free space toward a target—and \textit{fine-step} relative increments for delicate grasp alignment, contact adjustment, or recovery. At every physical control step, a lightweight local monitor evaluates task-conditioned physical criteria without invoking the large model. Nominal execution sustains ongoing \textit{large-step} batches to maximize decision reuse, whereas violated criteria or missing observations trigger immediate mid-batch interruption, returning structured diagnostic evidence that lets the model adjust local granularity, revise the active operation, or replan subsequent subtasks on demand.

This tightly coupled design resolves three structural bottlenecks in prior paradigms. First, whereas existing hierarchical systems rely on static or heuristic update schedules~\cite{Shi2025HiRO} that decouple decision timing from physical state, MATCH turns operation-specific criteria into event-driven scheduling signals. Second, instead of deferring verification until an action chunk or subtask concludes~\cite{Chen2026CanAS}—leaving execution blind to intra-batch perturbations—our monitor checks physical conditions at control-step resolution to halt error accumulation mid-batch. Third, unlike post-hoc failure analyzers~\cite{Liu2023REFLECTSR} or external runtime watchdogs~\cite{Zhou2024CodeasMonitorCV} that sit atop separate low-level policies and merely trigger binary replanning, MATCH unifies action synthesis and monitoring within a single generalist model—using step-indexed diagnostic evidence to jointly adapt action granularity (\textit{large-step} vs.\ \textit{fine-step}), horizon length, and recovery scope across local adjustments, operation revisions, and subtask replanning.

Our main contributions are as follows:
\begin{enumerate}
    \item \textbf{Closed-Loop Timescale Adaptation.} We formulate an adaptive multi-timescale coordination mechanism that translates control-step physical verification into dynamic scheduling signals for high-level model decision horizons.
    \item \textbf{The MATCH Framework.} We instantiate this mechanism in a unified generalist architecture that couples reusable subtask planning, multi-granularity control (\textit{large-step} waypoints and \textit{fine-step} increments), and executable condition monitoring to support immediate mid-batch interruption and scoped recovery.
    \item \textbf{Empirical Validation and Mechanistic Analysis.} Across six manipulation tasks on RoboDojo, MATCH raises the average task score from 0.338 to 0.683 while cutting action decision batches by 24.6\% even under lower reasoning effort, with systematic ablations confirming that these gains arise from closed-loop coordination rather than lower query frequency alone.
\end{enumerate}
\section{Related Work}
\label{sec:related}

\subsection{General Models and Embodied Harness for Robot Control}

The application of general-purpose multimodal models in robotics has progressed from high-level semantic task planning toward direct low-level action generation~\cite{Driess2023PaLMEAE,Brohan2023RT2VM}. Zhang et al.~\cite{Zhang2026AnUR} systematically evaluated GPT-6 Astra as a zero-shot robot policy on RoboDojo~\cite{Chen2026RoboDojoAU}, demonstrating strong generalization across manipulation tasks without robot-specific fine-tuning, while revealing persistent bottlenecks in fine-grained manipulation, dynamic control, and complex bimanual coordination. To bridge this gap, \emph{Astranomy}~\cite{yang2026astranomy} compared direct action generation, agentic harnesses, and model-supervised low-level policies, highlighting the sensitivity of manipulation performance to action representations, execution horizons, and historical context. Subsequent studies further translate model reasoning into executable behaviors via structured tools, spatial grounding, and execution feedback~\cite{Zhang2026HarnessVS,Si2026FewerTB,Wang2026PyRUAREALRE}. 

Nevertheless, when a general-purpose model directly serves as the action generator, its high inference latency makes static execution horizons unsustainable—demanding principled mechanisms to govern when the model should be queried and how long each decision should persist.

\subsection{Multi-Timescale Architectures for Robot Control}

To reconcile costly semantic reasoning with continuous physical control, hierarchical robotic systems typically decouple decisions across modules operating at fixed timescales~\cite{Shi2025HiRO,Bjorck2025GR00TNA,Sun2026UniFSUF,Cai2026InternW0AF}. \emph{Hi Robot} separates high-level instruction reasoning from low-level vision-language-action (VLA) execution~\cite{Shi2025HiRO}, while \emph{GR00T N1} pairs a vision-language module with a fast diffusion transformer for motor control~\cite{Bjorck2025GR00TNA}. Similarly, \emph{UniFS} updates hierarchical representations at progressively slower frequencies to reuse semantic context~\cite{Sun2026UniFSUF}, and the $\pi$ series~\cite{Black20240AV,Intelligence202505AV} steers high-frequency action policies with high-level semantic goals. 

However, these architectures presuppose a dedicated, domain-trained low-level policy running continuously at fixed frequencies. In contrast, when a general-purpose large model directly outputs robot actions without a learned low-level policy, decision timing cannot be statically partitioned; instead, the execution horizon itself must dynamically adapt to stage-specific control demands.

\subsection{Execution Monitoring and Adaptive Replanning}

Execution feedback is essential for coordinating high-level reasoning with physical control~\cite{Huang2022InnerME}. Early diagnostic frameworks like \emph{REFLECT}~\cite{Liu2023REFLECTSR} summarize multimodal histories to explain failures post hoc, whereas online supervisors intervene during rollouts. For instance, \emph{FastHarness}~\cite{Lin2026FastHarnessER} modulates GPT-6 Astra's audit frequency on a low-level VLA using prior experience, while Chen et al.~\cite{Chen2026CanAS} use a lightweight visual monitor to verify subtask progress at chunk boundaries. Other methods enforce continuous runtime checks: \emph{Code-as-Monitor}~\cite{Zhou2024CodeasMonitorCV} compiles VLM-generated constraints to detect anomalies and trigger replanning for an independent policy, while \emph{Auriga}~\cite{Wang2026AurigaTR} embeds generic pose-deviation checks into synthesized control scripts via code-as-policy.

\textbf{Positioning of MATCH.}
Existing feedback mechanisms address distinct aspects of closed-loop execution without jointly coordinating direct generalist control and step-level physical monitoring.
Post-hoc analyzers like \emph{REFLECT}~\cite{Liu2023REFLECTSR} enable retrospective correction, while \emph{FastHarness}~\cite{Lin2026FastHarnessER} and Chen et al.~\cite{Chen2026CanAS} supervise separate VLA policies at chunk boundaries rather than interrupting ongoing chunks.
Runtime monitors like \emph{Code-as-Monitor}~\cite{Zhou2024CodeasMonitorCV} verify constraints around existing policies without jointly adapting the horizon of direct generalist actions.
Meanwhile, \emph{Auriga}~\cite{Wang2026AurigaTR} supports execution-time verification and recovery via trajectory previews and movement-level tolerance checks, but relies on an execution harness with tool calls or synthesized programs rather than variable-horizon direct control.
In contrast, MATCH integrates subtask planning, variable-horizon direct control, and executable task-conditioned monitoring within one generalist model, using step-level physical feedback to interrupt action batches and trigger scoped recovery.

\begin{figure*}[!t]
    \centering
    \includegraphics[width=0.98\textwidth]{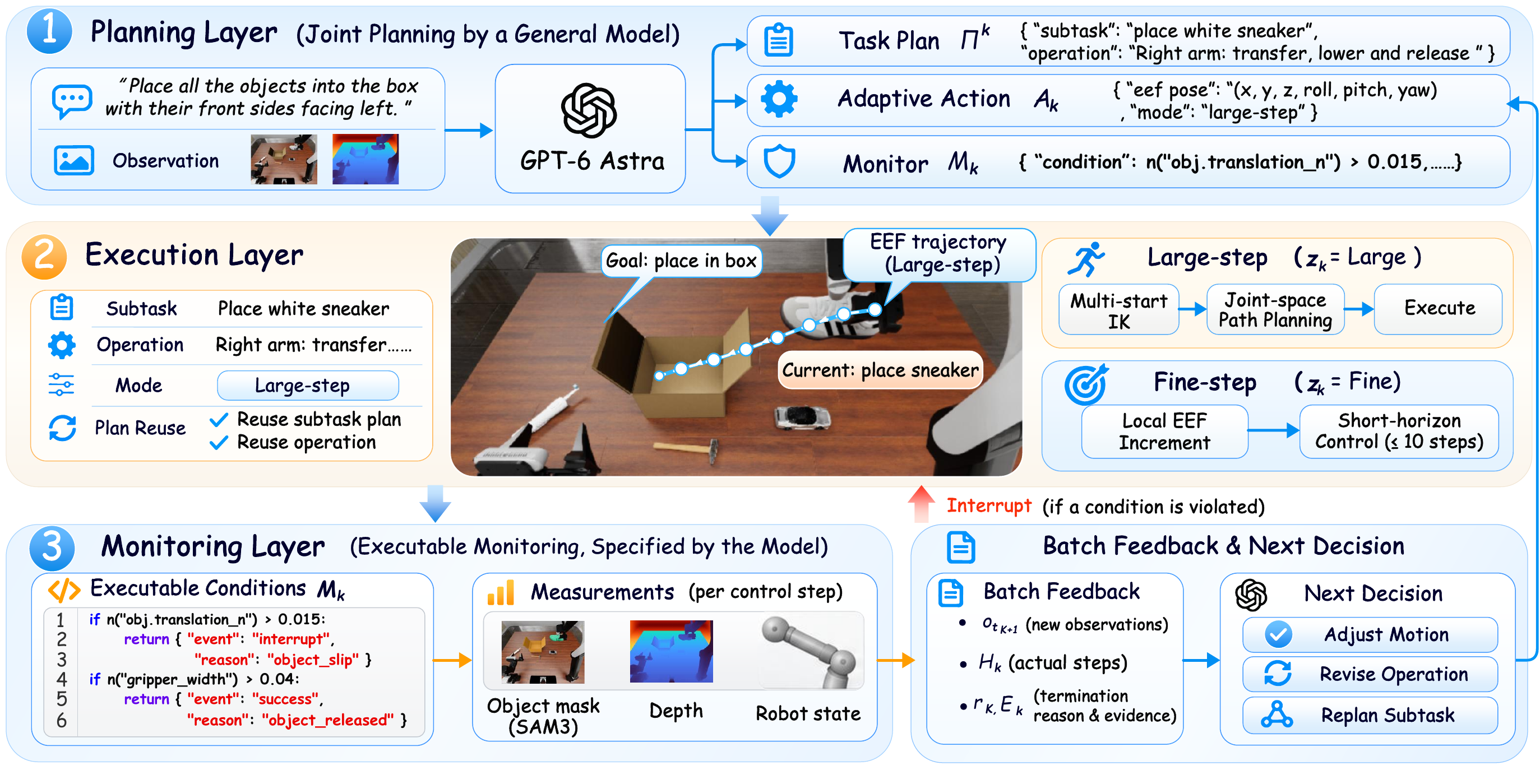}
    \caption{\textbf{Architecture of MATCH.} (1) \textbf{Planning Layer:} A single generalist model jointly generates reusable subtask plans, operation specifications, adaptive-horizon actions, and executable monitoring conditions. (2) \textbf{Execution Layer:} The local controller executes multi-step waypoints (\textit{large-step}) for extended motion or short-horizon increments (\textit{fine-step}) for precision alignment. (3) \textbf{Monitoring Layer:} Evaluating task conditions at every control step, the local monitor triggers mid-batch interruption upon violations and feeds structured evidence back to guide motion adjustment, operation revision, or subtask replanning.}
    \label{fig:architecture}
\end{figure*}
\section{Method}
\label{sec:method}

\subsection{Overview and Problem Formulation}
\label{sec:overview}

Given a natural-language task instruction $L$ and robot observation $o_t$, MATCH coordinates model decisions and physical execution through monitored action batches. A single general-purpose model $\pi_\theta$ generates task plans, operation specifications, actions, and monitoring specifications. A local controller executes end-effector (EEF) commands, while a local monitor evaluates the active monitoring conditions at each physical control step. We distinguish the control step $t$, the action decision index $k$, and the decision time $t_k$, measured in control steps.

The context available at the $k$-th decision is
\begin{equation}
    \mathcal{S}_k =
    \left(
    \Pi^{(k)}, i_k, \mathcal{O}_k,
    \mathcal{M}_k, h_k
    \right),
    \label{eq:hierarchical_context}
\end{equation}
where $\Pi^{(k)}$ is the task plan, $i_k$ indexes the active subtask, $\mathcal{O}_k$ specifies the current operation, $\mathcal{M}_k$ is the monitoring specification, and $h_k$ contains historical execution feedback.

MATCH operates at three timescales: task-level context persists across batches, actions are generated at batch boundaries, and monitoring conditions are evaluated at control-step resolution. The next decision time depends on both the proposed execution horizon and the observed execution state. Algorithm~\ref{alg:match} summarizes this closed-loop execution and selective update process. Its \textsc{Initialize} and \textsc{Coordinate} routines use the same model $\pi_\theta$: the former constructs the initial context and action, whereas the latter processes feedback to return either task termination or an updated context and next action.

\subsection{Task-Level Planning and Operation Specification}

The model decomposes the task into ordered subtasks. The active subtask in $\Pi^{(k)}$ is represented as
\begin{equation}
    \Pi_{i_k}^{(k)} = (g_k, \sigma_k, c_k),
    \label{eq:subtask}
\end{equation}
where $g_k$ denotes the subtask goal, $\sigma_k$ the operation strategy, and $c_k$ the high-level execution precautions. The current operation $\mathcal{O}_k$ specifies the manipulated object, selected arm, approach direction, and grasping or movement method. While the plan establishes what to achieve and in what order, the operation specifies the physical interaction used to advance that plan, and $c_k$ is subsequently grounded into executable monitoring specifications $\mathcal{M}_k$ for the active operation.

This task-level context is distinct from the geometric commands in an action batch. For example, a placement subtask defines the object and destination, its operation specifies the arm and transport strategy, while EEF waypoints are generated from observed states. Several batches can therefore pursue the same goal using different geometric commands.

At initialization or replanning, the model jointly generates the relevant plan, operation specification, monitoring specification, and first action batch. Subsequent batches reuse the applicable task-level context, with revisions handled by the feedback-driven coordination step. Retaining unaffected components and emitting only selective updates alongside each batch avoids regenerating full specifications, keeping per-call output token overhead low.

\subsection{Adaptive Action Generation and Execution}

Within a single model invocation (\textsc{Initialize} or \textsc{Coordinate}), conditioned on the active subtask selected from $\mathcal{S}_k$ and its associated operation, monitoring specification, and feedback history, the model emits any selective context updates jointly with the action batch:
\begin{equation}
\begin{aligned}
    A_k &= (W_k,z_k,H_k) \\
        &= \pi_\theta\bigl(
            L,o_{t_k},\Pi_{i_k}^{(k)},
            \mathcal{O}_k,\mathcal{M}_k,h_k
            \bigr),
\end{aligned}
\label{eq:action}
\end{equation}
where $W_k$ contains motion and gripper commands, $z_k\in\{\mathrm{Large},\mathrm{Fine}\}$ selects the execution mode, and $H_k\in\mathbb{Z}_{>0}$ bounds the number of control steps. Here, $\Pi_{i_k}^{(k)}$ selects the active subtask from Equation~\ref{eq:hierarchical_context}; the full plan remains available to \textsc{Coordinate} for task progression.

The modes differ in both action representation and execution horizon. In large-step mode, $W_k$ contains EEF pose waypoints followed by the local controller over multiple steps, supporting extended execution during free-space motion or stable transport. In fine-step mode, $W_k$ contains relative EEF displacements, supporting fine-grained corrections during grasp alignment, contact adjustment, or recovery. The mode determines how motion is represented, whereas the horizon limits how long the commands may be executed before another model decision.

The model selects both the mode and horizon according to the current operation and execution feedback. Large-step horizons are limited by the remaining task and subtask budgets. Fine-step batches execute at most 10 control steps ($H_k\leq 10$) and obey the applicable recovery requirements. In either mode, the proposed horizon is an upper bound, not a commitment to reach the commanded endpoint. Subsequent actions are conditioned on the state actually reached; the mid-batch early-interruption mechanism is defined below.

\begin{algorithm}[t]
\caption{MATCH: Monitored Execution and Selective Updates}
\label{alg:match}
\small
\begin{algorithmic}[1]
\Require Task instruction $L$, initial observation $o_0$
\State $k \gets 0,\quad t_0 \gets 0$
\State $(\mathcal{S}_0,A_0) \gets \Call{Initialize}{L,o_0}$

\While{\textbf{true}}
    \State $(W_k,z_k,H_k) \gets A_k$
    \State $E_k \gets [\,],\quad r_k \gets \mathrm{horizon}$

    \For{$s=1,\ldots,H_k$}
        \State $o_{t_k+s} \gets \Call{ControllerStep}{W_k,z_k}$
        \State $(m,e) \gets \Call{Evaluate}{\mathcal{M}_k,o_{t_k+s}}$
        \State $E_k \gets \Call{Append}{E_k,(s,m,e)}$

        \If{$z_k=\mathrm{Large} \land m\neq\mathrm{normal}$}
            \State $r_k \gets m$
            \State \textbf{break} \Comment{Monitor intervention}
        \ElsIf{\Call{ExecutionStop}{}}
            \State $r_k \gets \mathrm{execution\_stop}$
            \State \textbf{break}
        \EndIf
    \EndFor

    \State $\hat{H}_k \gets s,\quad t_{k+1} \gets t_k+\hat{H}_k$
    \State $f_k \gets (\hat{H}_k,r_k,E_k)$
    \State $(d,\mathcal{S}_{k+1},A_{k+1}) \gets$
    \Statex \hspace{\algorithmicindent}%
        $\Call{Coordinate}{L,o_{t_{k+1}},\mathcal{S}_k,f_k}$

    \If{$d=\mathrm{terminate}$}
        \State \textbf{break}
    \EndIf
    \State $k \gets k+1$
\EndWhile
\end{algorithmic}
\end{algorithm}

\subsection{Task-Conditioned Monitoring and Adaptive Intervention}

For the current operation, the model generates a monitoring specification $\mathcal{M}_k$ identifying the measured objects, anomaly criteria, and triggering reasons. The monitoring conditions are restricted Python Boolean expressions over current, reference, and historical values of registered measurements. The system validates these expressions and their measurement references before use. Perception and robot-state interfaces supply the measurement values, and the monitor evaluates the conditions locally at each control step without invoking the general-purpose model.

For example, during transport, potential object slippage can be detected using the condition $\Delta d_{\mathrm{og}} > \delta_{\mathrm{slip}}$. Here, $\Delta d_{\mathrm{og}}$ denotes the change in object position relative to the gripper from its registered reference, and $\delta_{\mathrm{slip}}$ is the corresponding threshold. This condition should be revised before intentional release, when relative object motion is expected. Monitoring specifications and reference states are reused while their measurement definitions and operational meaning remain applicable; revised specifications are revalidated before use. Reusing a specification also preserves its reference state, rather than resetting the comparison baseline at every action decision.

At each step, the monitor returns a status $m$ and evidence $e$. The status is $\mathrm{normal}$ when no violation is detected, $\mathrm{triggered}$ when an interruption criterion is satisfied, or $\mathrm{unavailable}$ when a required measurement cannot be obtained. For the transport example, evidence can include the measured displacement and triggering threshold; an unavailable outcome instead identifies a missing measurement. Missing measurements are therefore not treated as evidence of normal execution. In large-step mode, either non-normal outcome interrupts the remaining actions and returns its interruption reason and measurement evidence. Triggered conditions activate the applicable recovery requirements. For unavailable measurements, the recovery rule calls for fine-step execution until the required measurements are restored. In fine-step mode, monitoring outcomes are recorded and returned at the batch boundary without directly interrupting the current batch.

For an executable batch, let $\bar{H}_k\in\{1,\ldots,H_k\}$ denote the uninterrupted execution length---that is, the number of steps before reaching $H_k$ or satisfying \textsc{ExecutionStop} in the absence of task-monitoring interruption. In Algorithm~\ref{alg:match}, \textsc{ExecutionStop} covers target attainment, budget exhaustion, or a local-controller stop, and the label $\mathrm{execution\_stop}$ groups these three outcomes to distinguish them from monitoring intervention and horizon completion. Let $\tau_k$ be the first local step with a non-normal monitoring outcome, with $\tau_k=\infty$ if none is observed. The actual execution length $\hat{H}_k$ and next decision time $t_{k+1}$ satisfy
\begin{equation}
\begin{aligned}
\hat{H}_k &=
\begin{cases}
    \min\{\bar{H}_k,\tau_k\}, & z_k=\mathrm{Large},\\[4pt]
    \bar{H}_k, & z_k=\mathrm{Fine},
\end{cases}\\
t_{k+1} &= t_k+\hat{H}_k.
\end{aligned}
\label{eq:horizon}
\end{equation}
This asymmetric interruption rule is intentional: because fine-step batches already have short horizons ($H_k\leq 10$) and operate during grasp alignment, contact adjustment, or recovery where transient physical fluctuations are common, completing the batch yields a full step-indexed trace $E_k$ that helps distinguish momentary contact noise from persistent failure without triggering excessive model queries.

\subsection{Feedback-Driven Coordination and Task Progression}

At each batch boundary, MATCH returns the reached observation $o_{t_{k+1}}$ and feedback $f_k=(\hat{H}_k,r_k,E_k)$, comprising the actual execution length, batch termination reason, and monitoring evidence. The ordered sequence $E_k$ contains step-indexed records $(s,m,e)$, preserving the timing and recurrence of monitoring outcomes. \textsc{Coordinate} uses this feedback together with $\mathcal{S}_k$ to determine the next update and incorporate the execution evidence into the history $h_{k+1}$.

Updates have three scopes: correcting motion commands, modes, or horizons; revising the current operation and its monitoring specification; or replanning affected subtasks when their strategy, goals, or ordering are no longer applicable. Aside from mandatory safety overrides such as restoring unavailable measurements, the model selects the recovery scope from observations and execution history rather than a fixed mapping from monitoring events to replanning actions, while retaining unaffected task-level context.

Subtask completion is determined by the model, not automatically by a monitoring event or batch interruption. When switching subtasks, it records completion, failure, or deferral together with supporting evidence, then issues the next subtask context and first action batch. If \textsc{Coordinate} returns task termination, no subsequent action is executed.
\section{Experiments}
\label{sec:experiments}

\subsection{Experimental Setup and Research Questions}

\begin{figure}[t]
    \centering
    \includegraphics[width=\columnwidth]{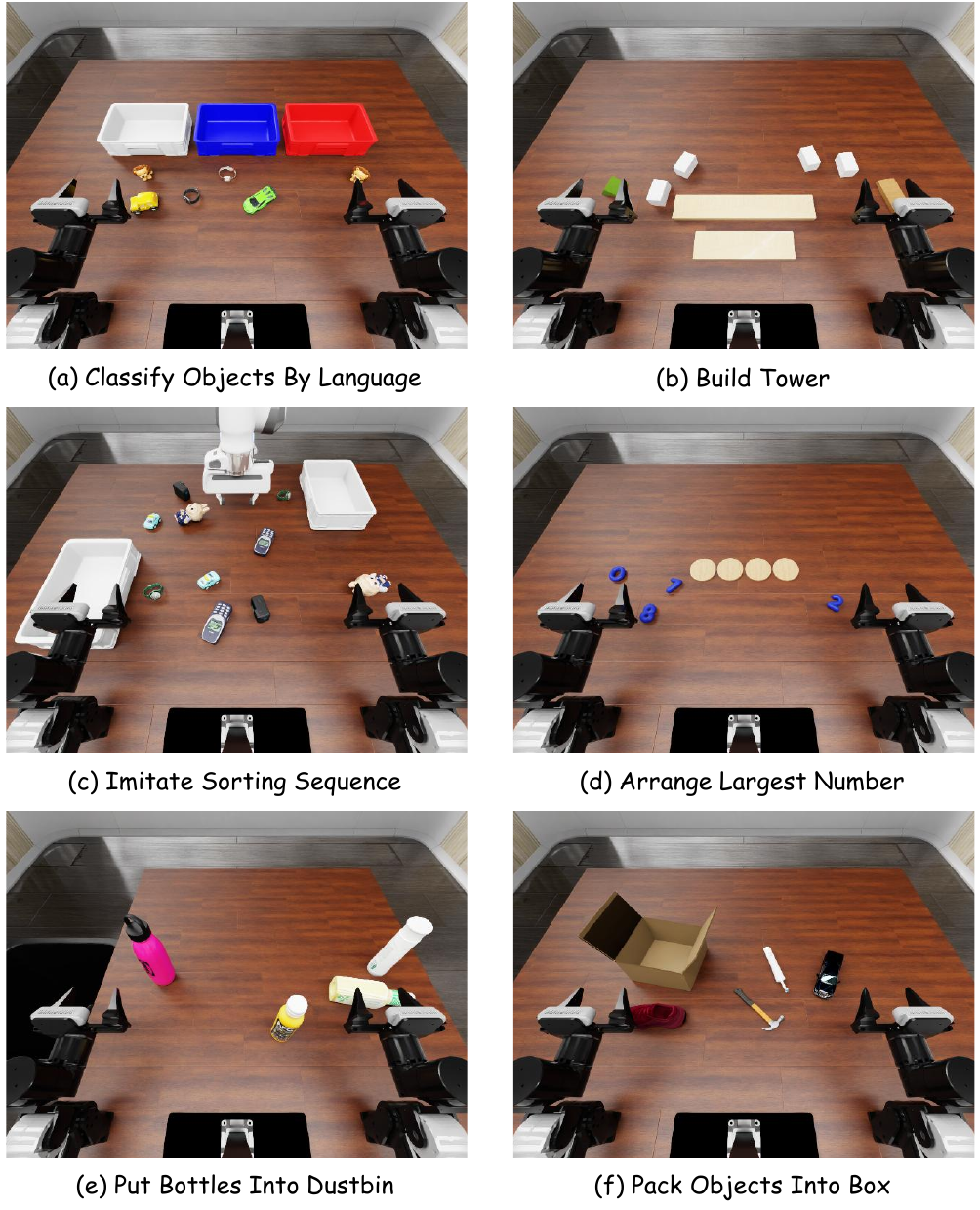}
    \caption{\textbf{Overview of the six RoboDojo manipulation tasks.} Evaluated on a dual-arm tabletop platform, these tasks cover all five official RoboDojo capability categories: \textit{Generalization}, \textit{Memory}, \textit{Precision}, \textit{Long-Horizon}, and \textit{Open}.}
    \label{fig:tasks}
\end{figure}

We evaluate MATCH across six manipulation task categories on the RoboDojo simulation benchmark~\cite{Chen2026RoboDojoAU} (Figure~\ref{fig:tasks}): \textit{Classify Objects By Language}, \textit{Build Tower}, \textit{Imitate Sorting Sequence}, \textit{Arrange Largest Number}, \textit{Put Bottles Into Dustbin}, and \textit{Pack Objects Into Box}. Following standard evaluation protocols, each task is tested across five distinct layouts under Seed 0.

Since generalist model inference dominates overall latency in physical robot control, we evaluate system efficiency by isolating algorithmic decision frequency from confounding infrastructure factors such as network fluctuation and server load. Specifically, we adopt two primary metrics: \textbf{task score (Score)} and \textbf{action decision batches (Action Batches)}. Score denotes the environment-returned task completion score in $[0, 1]$, where higher values indicate superior manipulation performance. Action Batches measures the average number of high-level model queries per episode. Because local condition monitoring operates at control-step frequency with negligible overhead relative to generalist model inference, fewer action batches directly translate to lower model interaction cost and higher decision efficiency.

Our evaluation investigates two core research questions:
\begin{itemize}
    \item \textbf{RQ1 (Effectiveness \& Efficiency):} Can closed-loop multi-timescale coordination simultaneously improve task completion performance and reduce high-level model decision frequency?
    \item \textbf{RQ2 (Mechanism Attribution):} Do performance gains stem from the closed-loop synergy of adaptive horizons, task planning, and runtime monitoring, rather than merely lowering decision frequency?
\end{itemize}

\subsection{RQ1: Main Results}

We benchmark MATCH against baselines spanning two execution paradigms on RoboDojo. Our primary direct-control baseline is \textbf{GPT-as-Policy (GAP Direct)}~\cite{su2026astra}, which directly generates robot actions from a generalist model using \texttt{XHigh} reasoning effort. To contextualize overall task difficulty, we also include two tool-augmented reference methods: \textbf{PhysicalRSI}~\cite{MMLab_PhysicalRSI_contributors_Physical_RSI_1_0_2026}, which pairs a VLA model with discrete manipulation tools under \texttt{XHigh} reasoning effort, and \textbf{RoboDawn}~\cite{Guo2026TransferringTI}, which leverages discrete tools alongside a single expert demonstration under \texttt{High} reasoning effort. Notably, MATCH operates purely in a zero-shot, demonstration-free setting using only \texttt{Low} reasoning effort.

\begin{table*}[!t]
    \centering
    \caption{\textbf{Task performance comparison on RoboDojo.}
    Score denotes the average task completion score in $[0, 1]$ across five layouts per task; bold indicates the best result. Methods are grouped into tool/demonstration-assisted references and direct generalist policy control.}
    \label{tab:task_performance}
    \small
    \setlength{\tabcolsep}{6pt}
    \renewcommand{\arraystretch}{1.12}
    \begin{tabular*}{\textwidth}{@{\extracolsep{\fill}}lcc|cc@{}}
        \toprule
        & \multicolumn{2}{c|}{\textit{Tool / Demo-Assisted References}} & \multicolumn{2}{c}{\textit{Direct Generalist Control}} \\
        \cmidrule(lr){2-3} \cmidrule(lr){4-5}
        Task
        & \makecell{PhysicalRSI\\(XHigh)}
        & \makecell{RoboDawn\\(High, 1-Demo)}
        & \makecell{GAP Direct\\(XHigh, Zero-Shot)}
        & \makecell{MATCH\\(Low, Zero-Shot)} \\
        \midrule
        Classify Objects By Language
        & 0.10 & 0.44 & 0.48 & \textbf{0.68} \\
        Build Tower
        & 0.25 & \textbf{0.54} & 0.12 & 0.36 \\
        Imitate Sorting Sequence
        & 0.00 & 0.66 & 0.00 & \textbf{0.83} \\
        Arrange Largest Number
        & 0.32 & \textbf{1.00} & 0.57 & \textbf{1.00} \\
        Put Bottles Into Dustbin
        & 0.46 & \textbf{0.88} & 0.36 & 0.73 \\
        Pack Objects Into Box
        & 0.14 & 0.19 & \textbf{0.50} & \textbf{0.50} \\
        \midrule
        \textbf{Overall (Mean)}
        & 0.212 & 0.618 & 0.338 & \textbf{0.683} \\
        \bottomrule
    \end{tabular*}
\end{table*}

\noindent\textbf{Task Performance.}
As shown in Table~\ref{tab:task_performance}, MATCH achieves an overall mean score of \textbf{0.683}, outperforming both the direct-control baseline GAP Direct (0.338) and the demonstration-assisted reference RoboDawn (0.618). Compared with GAP Direct, MATCH boosts the average score by \textbf{34.5 percentage points} while lowering model reasoning effort from \texttt{XHigh} to \texttt{Low}. Furthermore, despite requiring neither task-specific demonstrations nor pre-defined discrete manipulation skills, MATCH matches or surpasses RoboDawn on four of the six task categories—including a notable margin on \textit{Pack Objects Into Box} (0.50 vs.\ 0.19)—and attains the highest overall score. This confirms that principled closed-loop coordination between high-level reasoning and low-level physical execution outweighs brute-force inference scaling.

\noindent\textbf{Action Decision Efficiency.}
To assess interaction efficiency within the direct-control paradigm, Table~\ref{tab:batches} compares the average action decision batches between MATCH and GAP Direct. MATCH reduces model queries across five of the six task categories—achieving up to a $62.2\%$ reduction on \textit{Arrange Largest Number} and $48.3\%$ on \textit{Put Bottles Into Dustbin}—and lowers the overall mean from 167.1 to \textbf{126.0} batches ($-24.6\%$).

In \textit{Imitate Sorting Sequence}, GAP Direct terminates prematurely due to unrecovered early execution failures (scoring 0.00 across 44.0 batches), resulting in an artificially low batch count. By contrast, MATCH detects and recovers from intermediate anomalies via local monitoring, sustaining long-horizon multi-stage execution to achieve a \textbf{0.83} task score across 140.8 batches.

\par\vspace{6pt}
\noindent\begin{minipage}{\columnwidth}
    \centering
    \captionof{table}{\textbf{Action decision efficiency comparison under direct generalist control.}
    Batches ($\downarrow$) denotes the average number of model queries per episode.}
    \label{tab:batches}
    \footnotesize
    \setlength{\tabcolsep}{4.5pt}
    \renewcommand{\arraystretch}{1.15}
    \begin{tabular}{lrrr}
        \toprule
        Task
        & \makecell[r]{GAP Direct\\(XHigh) $\downarrow$}
        & \makecell[r]{MATCH\\(Low) $\downarrow$}
        & \makecell[r]{Relative\\Change} \\
        \midrule
        Classify Objects By Language  & 182.8 & \textbf{155.0} & $-15.2\%$ \\
        Build Tower                   & 210.2 & \textbf{146.0} & $-30.5\%$ \\
        Imitate Sorting Sequence      & 44.0\rlap{$^\dagger$} & 140.8 & $+220.0\%$ \\
        Arrange Largest Number        & 190.6 & \textbf{72.0}  & $-62.2\%$ \\
        Put Bottles Into Dustbin      & 136.2 & \textbf{70.4}  & $-48.3\%$ \\
        Pack Objects Into Box         & 238.8 & \textbf{171.8} & $-28.1\%$ \\
        \midrule
        \textbf{Overall (Mean)}       & 167.1 & \textbf{126.0} & $\mathbf{-24.6\%}$ \\
        \bottomrule
        \multicolumn{4}{l}{\scriptsize $^\dagger$Premature termination due to early failure (0.00 score).}
    \end{tabular}
\end{minipage}
\par\vspace{6pt}

\FloatBarrier

\begin{table*}[!t]
    \centering
    \caption{\textbf{Component ablation across two representative manipulation regimes.}
    All variants are evaluated with GPT-6 Astra (\texttt{Low}) across five layouts per task. Bold indicates the best task score.}
    \label{tab:ablation}
    \small
    \setlength{\tabcolsep}{6pt}
    \renewcommand{\arraystretch}{1.12}
    \begin{tabular*}{\textwidth}{@{\extracolsep{\fill}}l cc | cc | cc@{}}
        \toprule
        & \multicolumn{2}{c|}{\textit{Put Bottles Into Dustbin}}
        & \multicolumn{2}{c|}{\textit{Build Tower}}
        & \multicolumn{2}{c}{\textit{Mean}} \\
        \cmidrule(lr){2-3} \cmidrule(lr){4-5} \cmidrule(lr){6-7}
        Variant
        & Score $\uparrow$ & Batches $\downarrow$
        & Score $\uparrow$ & Batches $\downarrow$
        & Score $\uparrow$ & Batches $\downarrow$ \\
        \midrule
        GAP (Fixed Step = 10)
        & 0.34 & 69.6
        & 0.08 & 126.2
        & 0.210 & 97.9 \\
        MATCH (Without Planning)
        & 0.46 & 86.6
        & 0.08 & 138.8
        & 0.270 & 112.7 \\
        MATCH (Without Monitoring)
        & 0.61 & 62.0
        & 0.22 & 110.4
        & 0.415 & 86.2 \\
        MATCH (Full)
        & \textbf{0.73} & 70.4
        & \textbf{0.36} & 146.0
        & \textbf{0.545} & 108.2 \\
        \bottomrule
    \end{tabular*}
\end{table*}

\subsection{RQ2: Ablation and Multi-Timescale Dynamics}

We conduct ablations across two physical interaction regimes: \textit{Put Bottles Into Dustbin}, pairing extended free-space transport with slip-prone grasping, and \textit{Build Tower}, requiring high-precision sequential alignment where placement errors compound across layers. All variants use GPT-6 Astra (\texttt{Low}) to eliminate reasoning-effort variance:
\begin{itemize}
    \item \textbf{GAP (Fixed Step = 10):} Doubles GAP Direct's fixed decision interval from 5 to 10 steps, testing static horizon extension alone.
    \item \textbf{MATCH (Without Planning):} Retains adaptive horizons and local monitoring but removes subtask decomposition and operation specifications.
    \item \textbf{MATCH (Without Monitoring):} Retains task planning and multi-granularity action generation but disables mid-batch monitoring interruptions.
    \item \textbf{MATCH (Full):} The closed-loop framework integrating all three mechanisms.
\end{itemize}

\noindent\textbf{Adaptive Horizons and Task Planning vs.\ Fixed Decision Intervals.}
Comparing fixed-horizon baselines against MATCH variants exposes an efficiency--precision dilemma under static intervals. With a short fixed step (\textbf{GAP Direct}, step $=5$), frequent re-querying during free-space transit incurs overhead ($136.2$ batches on \textit{Put Bottles Into Dustbin} and $210.2$ on \textit{Build Tower}), while decisions lacking hierarchical constraints or multi-granularity actions yield low scores ($0.36$ and $0.12$). Conversely, doubling the chunk in \textbf{GAP (Fixed Step = 10)} matches MATCH's query budget on \textit{Put Bottles Into Dustbin} ($69.6$ vs.\ $70.4$ batches), yet a 10-step horizon induces intra-batch open-loop blindness during alignment and placement—preventing step shortening or mid-batch braking, stalling the score at $0.34$ and collapsing \textit{Build Tower} to $0.08$.

By contrast, coupling reusable planning with variable-horizon multi-granularity actions resolves this trade-off, with both proving essential. Removing planning (\textbf{MATCH (Without Planning)}) removes persistent cross-batch context; even with adaptive horizons and monitoring, repeated goal re-inference inflates decision batches ($86.6$ on \textit{Put Bottles Into Dustbin}, $+23.0\%$ over Full, and $138.8$ on \textit{Build Tower}) while capping scores at $0.46$ and $0.08$. Conversely, retaining reusable plans alongside multi-granularity synthesis (\textbf{MATCH (Without Monitoring)}) lets the controller assign \textit{large-step} pose waypoints to free-space transit and \textit{fine-step} increments to contact alignment—lifting \textit{Put Bottles Into Dustbin} to \textbf{0.61} ($62.0$ batches) and boosting \textit{Build Tower} from $0.08$ to \textbf{0.22} ($110.4$ batches) without runtime interruptions, unlocking multi-layer stacking.

\noindent\textbf{Multi-Granularity Mode Switching and Mid-Batch Horizon Truncation.}
Layering physical monitoring atop variable-horizon planning elevates batch-level adaptability into control-step closed-loop execution. On \textit{Put Bottles Into Dustbin}, mid-batch monitoring catches slip-prone failures early, raising the score from $0.61$ to \textbf{0.73} ($70.4$ batches). On contact-sensitive \textit{Build Tower}, \textbf{MATCH (Full)} adds a \textbf{+0.14} gain (\textbf{+63.6\%} relative improvement) over \textbf{MATCH (Without Monitoring)} to reach \textbf{0.36} ($4.5\times$ the static-chunk baseline): during upper-layer stacking, chunk-boundary granularity alone cannot absorb sudden contact perturbations, whereas control-step monitoring halts misaligned motions mid-batch before toppling lower blocks—with the extra queries ($146.0$ vs.\ $110.4$ batches) reflecting productive recovery into higher layers while still cutting queries by $30.5\%$ below GAP Direct ($210.2$ batches).

Execution logs across five \textbf{MATCH (Full)} episodes on \textit{Put Bottles Into Dustbin} ($352$ batches) quantify how monitoring drives multi-granularity scheduling. The model allocates $44.0\%$ ($155$) of batches to \textit{large-step} pose waypoints for rapid free-space traversal and $56.0\%$ ($197$) to \textit{fine-step} relative increments ($H_k \leq 10$) for precision alignment. Crucially, \textbf{41 of the 155 \textit{large-step} batches (26.5\%)} are truncated early mid-batch ($\hat{H}_k < H_k$) upon monitoring triggers, prompting immediate switches to \textit{fine-step} recovery in \textbf{28 cases (68.3\%)} and revised \textit{large-step} resumption in \textbf{13 cases (31.7\%)}. These statistics confirm runtime monitoring acts as an active closed-loop scheduler governing multi-granularity adaptation rather than a passive failure alarm.
\section{Conclusion}
\label{sec:conclusion}

We propose MATCH, an adaptive multi-timescale framework for generalist robot manipulation. Combining task planning, multi-granularity execution, and task-conditioned monitoring, MATCH coordinates high-level reasoning with low-level control, reusing decisions in stable stages while adapting to fine contact or anomalies for balanced performance and efficiency.

On RoboDojo, MATCH improves the average task score by \textbf{34.5 percentage points} and reduces action batches by \textbf{24.6\%} over fixed-frequency direct control. Ablations further validate the synergy of planning and monitoring.

While evaluated on the RoboDojo simulation benchmark, extending MATCH to physical robots remains future work to validate our multi-timescale framework under real-world dynamics and real-time operational constraints.

\section*{LLM Usage Statement}
During the preparation of this work, large language models (LLMs), including OpenAI's GPT models, and generative AI tools were used strictly in an assistive capacity for: (1) coding assistance, debugging, and experimental script refactoring; (2) manuscript writing support, including language polishing, grammar checking, and clarity improvement; (3) \LaTeX{} formatting and table/algorithm typesetting; and (4) drafting and refining illustrative figures and diagrams. All core research ideas, framework designs, experimental evaluations, data analyses, and scientific claims were conceived, verified, and authored by the human authors, who take full responsibility for the content of this paper.

{
    \small
    \bibliographystyle{ieeenat_fullname}
    \bibliography{main}
}

\clearpage
\MATCHAppendixFloats
\raggedbottom
\appendix
\makeatletter
\def\@title{RoboMATCH: Monitoring-Augmented Task and Control Hierarchy \\
for Adaptive Multi-Timescale Coordination in Generalist Robot Manipulation \\[0.5em]
\large Supplementary Material}
\def\@author{Zihan Yang et al.}
\twocolumn[\MATCHOriginalTitle]
\makeatother
\renewcommand{\thetable}{S\arabic{table}}
\renewcommand{\thefigure}{S\arabic{figure}}
\setcounter{table}{0}
\setcounter{figure}{0}

This supplementary material provides implementation details, runtime analyses, and additional experimental evidence supporting MATCH:
\begin{itemize}
  \item \textbf{Sections~A--C} detail the model input/prompt structure, hierarchical planning-to-action grounding, task-conditioned monitoring interface, and perception/low-level control mechanisms.
  \item \textbf{Sections~D--F} specify the RoboDojo evaluation protocol, baseline configurations, per-layout results across six tasks, and full ablation and sensitivity analyses.
  \item \textbf{Sections~G--H} present runtime, latency, and token measurements alongside representative mid-batch intervention traces, interruption statistics, and failure attribution.
\end{itemize}

\section{Implementation and Prompt Details}
\label{sec:impl_prompt}

This section supplements the system implementation and prompt design in Sections~3--4 of the main paper. Following the closed-loop cycle---``task instruction and initial observation $\rightarrow$ joint planning and action submission $\rightarrow$ runtime validation and execution $\rightarrow$ feedback-driven next decision''---we detail how MATCH progressively grounds high-level task plans into robot end-effector actions and returns control to the general-purpose model when needed.

\subsection{Experimental Configuration and Model Inputs}

Consistent with Sections~4.2--4.3 of the main paper, MATCH uses \textbf{GPT-6 Astra (Low)} as the sole general-purpose decision model across all main experiments and ablation variants (Table~3), operating zero-shot without demonstrations. Within an accepted subtask, consecutive decisions share a continuous dialogue context. Upon switching to a new subtask, a fresh session is initialized; the runtime carries over only the current physical state, task progress, concise preceding evidence, and active monitoring/protection states. This subtask-scoped session boundary prevents unbounded visual-token accumulation and attention dilution over long-horizon rollouts (often exceeding 100 decision turns), while preserving exact physical continuity through structured state handoff.

At every decision turn, the MATCH runtime supplies four decoupled input channels (Table~\ref{tab:input_channels}), separating static control rules, dynamic physical feedback, raw visual observations, and structured output constraints. All model outputs are strictly validated against the tool schema before physical actuation, rejecting malformed parameters in place to guarantee execution safety and reproducibility.

\begin{table*}[t]
\centering
\footnotesize
\renewcommand{\arraystretch}{1.08}
\setlength{\tabcolsep}{5pt}
\caption{Four decoupled input channels provided to the decision model at each turn.}
\label{tab:input_channels}
\begin{adjustbox}{max width=\linewidth}
\begin{tabular}{@{}L{0.15\textwidth}L{0.20\textwidth}L{0.34\textwidth}L{0.25\textwidth}@{}}
\toprule
\textbf{Input channel} & \textbf{Update timing} & \textbf{Core contents} & \textbf{Boundary \& constraints} \\
\midrule
\textbf{System instruction} & Fixed at subtask session initialization & Control modes, large-/fine-step rules, planning--operation--action contract, monitoring rules, and visible task instruction & Restricted to the visible task contract; no hidden poses, contact ground truth, or rewards \\
\textbf{Dynamic observation JSON} & Updated every decision turn & Physical step and remaining budget, robot/EEF state, prior-batch tracking and motion feedback, registered-object measurements, and alarm-recovery state & Contains only observed and executed feedback \\
\textbf{Raw camera images} & Updated every decision turn & Three raw RGB views (top-view, left-wrist, and right-wrist cameras) & Passed as an independent visual modality \\
\textbf{Tool schema} & Fixed at the interface layer & Parameter structures and type constraints for action submission, motion diagnostics, and task termination & Enforces structured outputs via tool calling \\
\bottomrule
\end{tabular}
\end{adjustbox}
\end{table*}

\subsection{Prompt Design and Ablation Implementation}

\paragraph{Core System-Prompt Rules.}
Box~A.1 excerpts the core system-prompt rules governing large-/fine-step mode selection and the hierarchical contract (\texttt{Mission} $\rightarrow$ \texttt{Subtask} $\rightarrow$ \texttt{Operate} $\rightarrow$ \texttt{Action}). Crucially, the mode-selection rules enforce evidence-grounded scheduling: they explicitly prohibit defaulting to conservative fine-step execution out of generic uncertainty or inertia from prior batches, ensuring that the controller promptly resumes large-step traversal once feedback-sensitive contact phases conclude. The full prompt additionally specifies cross-subtask occlusion analysis, action-conditioned risk review, and monitoring registration (Appendix~\ref{sec:monitoring_details}). Prompt- and interface-level modifications for the Table~3 ablation variants are consolidated in Table~\ref{tab:ablation_matrix} (Appendix~\ref{sec:ablation_matrix}).

\begin{table*}[t]
\centering
\begin{promptbox}[unbreakable]{Box A.1: Core Control Rules and Hierarchical Contract (MATCH System Prompt)}
\textbf{CONTROL MODES AND BASIC RULES}\\
- Normal: choose duration from current evidence and remaining native episode steps, with no fixed batch-step ceiling; request at least 25 cycles, or all remaining cycles if fewer than 25 remain.\\
- Fine: explicitly choose 1 to fine\_step\_limit cycles from current visual evidence.\\[4pt]
\textbf{CONTROL MODE SELECTION}\\
- Large-step: use for continuous motion along an evidence-supported route; resume it when the feedback-sensitive phase ends, without requiring every preparation pose to be exact.\\
- Small-step: required for the next batch after a response=notify condition triggers. Otherwise, use it when the next observation can materially change the immediate interaction or recovery decision, or when recovery requires it. Generic uncertainty, an old alarm, unfinished free-space rotation or the previous mode alone does not justify small-step control.\\[4pt]
\textbf{MISSION, SUBTASK, OPERATE AND ACTION CONTRACT}\\
The same control model publishes the first subtask, its operate and the first action together. Every later observation requires a newly reassessed operate and a new action submission; the current subtask remains fixed until a subtask switch is published.\\[3pt]
\textbf{OPERATE:} Operation goal is a verifiable local outcome; one operation may span several action batches. Before every action submission, publish plan.operate=\{operation\_id, subtask\_id, goal, observation\_step, arm\_reachability, interaction\_decisions\} using the current observation step. The same operation\_id may continue when the local method remains valid, but its interaction decisions must be refreshed from the latest evidence. Change operation\_id and provide reset when the interaction method or local goal changes.\\[3pt]
\textbf{ACTION:} Bind each action to the active subtask and the freshly published current operation\_id. Use the current operate to constrain every waypoint. The active EEF arm must appear in its action\_choice.
\end{promptbox}
\end{table*}

\paragraph{Turn-Level Decision Templates.}
During closed-loop execution, the runtime prompts the model with two user templates depending on whether the system is at a subtask boundary (Box~A.2). The initial turn (step~0) and post-switch planning turns use the initialization template; subsequent turns within the same subtask use the continuation template, requiring the model to reassess \texttt{operate} from the latest observation and jointly submit the next action. Both templates explicitly enforce single-turn action commitment---forbidding passive summarization turns without an executable tool call---so that every model invocation directly advances physical execution or declares an explicit blocker.

\begin{table*}[t]
\centering
\begin{promptbox}[unbreakable]{Box A.2: Turn-Level User Prompt Templates (Raw Camera Images Passed Separately)}
\textbf{[Initialization / New Planning Round (initial or next subtask)]}\\
Submit complete mission and subtask\_decision, a freshly reassessed operate for the current observation, and EEF waypoints together through submit\_action(mode, plan).\\
Current observation step: \{step\}; remaining native steps: \{remaining\_native\_steps\}. Apply the standing control rules.\\
Use the attached current execution result to choose and submit the next action in this turn. Do not stop after summarizing the previous action. If no safe evidence-supported action exists, explicitly state the blocker.\\
\{observation\_json\}\\[5pt]
\textbf{[Continue Current Subtask]}\\
Continue the current subtask. Before every action, reassess and publish operate for the current observation together with EEF actions; plan\_next\_subtask=true schedules the next planning turn. Record completed, failed or deferred with reason and evidence when switching.\\
Current observation step: \{step\}; remaining native steps: \{remaining\_native\_steps\}. Apply the standing control rules.\\
Use the attached current execution result to choose and submit the next action in this turn. Do not stop after summarizing the previous action. If no safe evidence-supported action exists, explicitly state the blocker.\\
\{current\_observation\_json\}
\end{promptbox}
\end{table*}

\subsection{From Planning to Actions: Hierarchical Binding and Physical Grounding}

Building on Sections~3.2--3.3 of the main paper, MATCH grounds abstract plans into concrete end-effector commands via explicit identifier binding and per-turn reassessment (Table~\ref{tab:hierarchy_binding}).

\begin{table*}[t]
\centering
\footnotesize
\renewcommand{\arraystretch}{1.08}
\setlength{\tabcolsep}{5pt}
\caption{Hierarchical information flow, identifier binding, and update lifecycles in MATCH.}
\label{tab:hierarchy_binding}
\begin{adjustbox}{max width=\linewidth}
\begin{tabular}{@{}L{0.14\textwidth}L{0.38\textwidth}L{0.42\textwidth}@{}}
\toprule
\textbf{Level} & \textbf{Core information carried} & \textbf{Binding and update lifecycle} \\
\midrule
\textbf{Subtask} & Subtask ID, goal, overall strategy, and cross-subtask occlusion dependencies & Immutable once published; remains active across batches until explicitly switched (\texttt{completed}, \texttt{failed}, or \texttt{deferred}) based on observational evidence \\
\textbf{Operate} & Operation ID, parent subtask, observation step, bimanual reachability, interaction decisions, and optional reset record & Bound to the active subtask; must be reassessed and republished after every new observation \\
\textbf{Action} & Execution mode, step limit, ordered EEF waypoints (position, quaternion, gripper opening, step granularity, grasped-object binding), and rationale & Strictly bound to the active subtask and current turn's operation ID; each submission drives one physical batch \\
\textbf{Monitor} & Segment ID, target, inline perception registration, Boolean abnormality predicates, and response level & Submitted jointly with the action; reused across batches when target, entities, and failure semantics are unchanged to preserve baselines, or replaced with a revision reason \\
\bottomrule
\end{tabular}
\end{adjustbox}
\end{table*}

\paragraph{First-Turn Joint Submission and Identifier Lifecycle.}
In the initial turn, the model jointly submits the mission plan, first subtask, initial operation specification, monitoring configuration, and first action batch in a single call. In subsequent batches, the active subtask remains fixed while the operation specification updates as follows:
\begin{itemize}
  \item \textbf{Reuse operation ID:} When the local goal and interaction method remain valid, the model retains the operation ID, updates the observation step, and refreshes interaction decisions without triggering a reset.
  \item \textbf{Replace operation ID:} When the local goal completes, fails, or requires a strategy change within the same subtask, the model publishes a new operation ID with a reset record (prior ID, outcome, and evidence); failures additionally specify the failure stage, cause, and corrective plan.
  \item \textbf{Subtask switching and replanning:} After logging the current subtask outcome from observational evidence, the model publishes the next subtask or revises the plan. Deferred or failed subtasks are rescheduled under their original IDs once dependencies are met.
\end{itemize}

\paragraph{Physical Grounding into Waypoints, Gripper Commands, and Monitors.}
Constraints propagate to execution through three mechanisms:
\begin{enumerate}
  \item \textbf{Arm and waypoint constraints:} The operation specification evaluates left-arm, right-arm, and bimanual reachability to select the preferred arm, then defines action choices (contact part, mode, approach direction, and release timing). The active EEF arm must match the current action choice (kinematics detailed in Section~3.3 of the main paper).
  \item \textbf{Gripper and swept-volume constraints:} Gripper closure is determined from visual finger alignment and proprioception; translational and rotational waypoints account for the swept collision volume of the open fingers.
  \item \textbf{Risk-to-monitor mapping:} Risks identified during operation specification (e.g., unintended lifting, slippage, tipping, or collision) are compiled prior to action submission into batch monitoring predicates via three risk reviews (object--object, robot--object, and object-state; Appendix~\ref{sec:monitoring_details}).
\end{enumerate}

\subsection{Counting Conventions for Batches and Model Calls}

To distinguish algorithmic decision frequency from low-level system overhead, Table~\ref{tab:counting_conventions} defines our two counting conventions.

\begin{table}[t]
\centering
\footnotesize
\renewcommand{\arraystretch}{1.08}
\setlength{\tabcolsep}{4pt}
\caption{Counting conventions for action batches and model decision turns.}
\label{tab:counting_conventions}
\begin{adjustbox}{max width=\linewidth}
\begin{tabular}{@{}L{0.28\columnwidth}L{0.68\columnwidth}@{}}
\toprule
\textbf{Convention} & \textbf{Definition} \\
\midrule
\textbf{Action Batches} & A group of executable actions validated and accepted once via the unified action interface (along with its planning update, operation reassessment, and monitoring rules), then executed continuously by the episode loop for multiple control steps before returning one execution receipt. \\
\textbf{Model Decision Turns} & A protocol-level decision turn initiated by the MATCH runtime, characterizing the total number of runtime--model interactions. \\
\bottomrule
\end{tabular}
\end{adjustbox}
\end{table}

Through a serialized closed-loop temporal protocol, the unified action interface guarantees strict one-to-one equality between these two quantities along normal execution paths:
\begin{itemize}
  \item \textbf{Single joint submission:} Planning updates, operation reassessment, actions, and monitoring rules are submitted together in one call; no separate planning or monitoring turns are opened.
  \item \textbf{Immediate turn closure:} Upon validating and accepting an action submission, the runtime closes the current decision turn before dispatching the batch, ensuring at most one accepted batch per model call.
  \item \textbf{Serialized execution and bundled feedback:} The execution loop runs one batch to completion, then bundles the execution receipt, updated observation, and next-step instruction into the subsequent model call without separate analysis turns.
  \item \textbf{Exception boundary:} Model decision turns strictly equal accepted action batches whenever calls are valid. Raw API calls slightly exceed accepted batches only when an occasional schema or parameter validation error triggers an in-place retry on the same observation.
\end{itemize}

\section{Monitoring Interface Details}
\label{sec:monitoring_details}

This section supplements Section~3.4 of the main paper with measurement variable definitions, restricted expression syntax, reference-value lifecycles, and runtime intervention mechanics. While Figure~2 of the main paper illustrates monitoring logic with intuitive pseudocode, runtime execution uniformly evaluates the restricted Boolean-expression interface defined below.

\subsection{Measurement Variables, Expression Syntax, and Units}

Monitoring predicates construct dynamic measurement keys from element aliases declared in the active segment (\texttt{A}, \texttt{B} denote registered visual targets or robot-body elements; \texttt{R} denotes a robotic arm). All spatial quantities use SI meters (m) and radians (rad). Table~\ref{tab:measurement_keys} lists all supported key families.

\begin{table*}[t]
\centering
\footnotesize
\renewcommand{\arraystretch}{1.08}
\setlength{\tabcolsep}{5pt}
\caption{Supported measurement-key families in the task-conditioned monitoring interface.}
\label{tab:measurement_keys}
\begin{adjustbox}{max width=\linewidth}
\begin{tabular}{@{}L{0.30\textwidth}L{0.28\textwidth}L{0.36\textwidth}@{}}
\toprule
\textbf{Measurement-key family} & \textbf{Meaning} & \textbf{Applicability and semantics} \\
\midrule
\texttt{A.x}, \texttt{A.y}, \texttt{A.z} & Element center in the environment frame & Registered visual targets and robot-body elements \\
\texttt{A.relative\_left.\{x,y,z\}}, \texttt{A.relative\_right.\{x,y,z\}} & Element center in the left/right EEF control frame & All elements relative to a measured EEF pose \\
\texttt{A.frame\_\{x,y,z\}.\{x,y,z\}} & Unit-axis components of the robot frame in the environment frame & Robotic-arm elements only \\
\texttt{A.axis.\{x,y,z\}}, \texttt{A.normal.\{x,y,z\}} & Fitted-line direction or plane normal components & Visual elements with a valid line/plane fit \\
\texttt{distance:A:B} & Euclidean distance between element centers & Any pair of available elements \\
\texttt{surface\_distance:A:B} & Geometry-supported surface distance & Pairs of a visible sampled surface and a nominal gripper pad/finger outer wall \\
\texttt{plane\_distance:A:B} & Distance from \texttt{A}'s center to \texttt{B}'s fitted plane & Available when \texttt{B} has a fitted plane normal \\
\texttt{A.sam3\_motion.\allowbreak\{dx,dy,dz,translation\_m\}}, \texttt{rotation\_rad}, \texttt{single\_view} & Object surface motion relative to a persistent world reference & Visual elements bound to a persistent world-motion reference (reliability reported independently) \\
\texttt{A.sam3\_grasp\_motion.\allowbreak\{dx,dy,dz,translation\_m\}}, \texttt{rotation\_rad}, \texttt{single\_view} & Surface motion in the gripper frame relative to a persistent grasp reference & Visual elements bound to an arm-specific grasp reference (\texttt{\{id, arm\}}) \\
\texttt{pred('distance:A:R')}, \texttt{pred('lower\_surface\_\allowbreak distance:A:R')}, \texttt{pred('commanded\_surface\_\allowbreak distance:R:A')} & Kinematic prediction for the next proposed step (\texttt{lower\_surface\_distance} subtracts swept motion; \texttt{commanded\_surface\_distance} evaluates at commanded pose) & Pre-step checks only (accessed via \texttt{pred()}); purely kinematics-based predictions \\
\bottomrule
\end{tabular}
\end{adjustbox}
\end{table*}

\paragraph{Expression Syntax.}
Each predicate is a restricted Python Boolean expression statically validated via AST parsing before installation. Allowed calls are restricted to measurement accessors (\texttt{m}, \texttt{ref}, \texttt{prev}, \texttt{pred}, \texttt{valid}) and numeric helpers (\texttt{abs}, \texttt{min}, \texttt{max}). Permitted operators include arithmetic (\texttt{+}, \texttt{-}, \texttt{*}, \texttt{/}), unary negation (\texttt{-}), comparisons (\texttt{<}, \texttt{<=}, \texttt{>}, \texttt{>=}, \texttt{==}, \texttt{!=}), and Boolean logic (\texttt{and}, \texttt{or}, \texttt{not}). Exponentiation (\texttt{**}), attribute access, indexing, loops, imports, and non-key string literals are prohibited, ensuring safe, low-latency step-level evaluation.

\subsection{Reference-Value Semantics and Monitoring-Segment Lifecycle}

\paragraph{Accessor Semantics.}
Within an installed monitoring segment, the five accessors operate as follows:
\begin{itemize}
  \item \texttt{m(k)}: Returns the finite, valid value of key \texttt{k} in the current evaluation frame.
  \item \texttt{ref(k)}: Latches the first finite value observed after segment installation as the baseline across steps and batches.
  \item \texttt{prev(k)}: Returns the value from the immediately preceding distinct evaluation frame.
  \item \texttt{pred(k)}: Returns the kinematic prediction for the next proposed action step.
  \item \texttt{valid(k)}: Checks whether key \texttt{k} has a valid measurement in the current frame.
\end{itemize}

\paragraph{Lifecycle and Reset Mechanism.}
Monitoring segments are stateful by identifier: submitting an identical ID and specification reuses the existing state and \texttt{ref()} baseline; submitting a new ID requires a \texttt{revision\_reason} and clears ordinary \texttt{ref()} and \texttt{prev()} caches. Opening the gripper automatically clears that arm's relative pose and grasp reference, whereas persistent SAM3 world and grasp references remain independently bound and unaffected by ordinary resets.

\subsection{Runtime Evaluation and Execution-Intervention Mechanism}

Following Section~3.4 and Algorithm~1 of the main paper, runtime intervention depends on the active control mode and evaluation state:
\begin{itemize}
  \item \textbf{Large-step mode:} The runtime installs or reuses the monitoring segment and evaluates every enabled predicate prior to each physical step and at batch end; any anomaly triggers immediate truncation.
  \item \textbf{Fine-step mode:} Within-batch monitoring interruption is disabled. Step-level evaluations are logged and summarized at the end of the short batch for model review, while low-level native safety checks remain active.
\end{itemize}

Table~\ref{tab:monitor_states} summarizes how each evaluation state affects Large-step execution and subsequent decisions. Table~\ref{tab:runtime_cases} presents four representative trajectories from the main experiments, illustrating distinct monitoring mechanisms and both typical recovery paths (Fine-step correction, or diagnosing a tracking-drift false positive, revising the monitor, and resuming Large-step). All thresholds are autonomously chosen by the model based on operation risk and physical units; the runtime contains no hard-coded cross-task thresholds.

\begin{table*}[t]
\centering
\footnotesize
\renewcommand{\arraystretch}{1.08}
\setlength{\tabcolsep}{5pt}
\caption{Effect of runtime monitoring states on Large-step execution and subsequent decisions.}
\label{tab:monitor_states}
\begin{adjustbox}{max width=\linewidth}
\begin{tabular}{@{}L{0.15\textwidth}L{0.33\textwidth}L{0.46\textwidth}@{}}
\toprule
\textbf{State} & \textbf{Trigger condition} & \textbf{Effect on current Large-step batch and next decision} \\
\midrule
\texttt{normal} & All required measurements are valid; no predicate triggers & Execution proceeds as planned until the target is reached or the step budget is exhausted \\
\texttt{triggered} & At least one abnormality predicate evaluates to true & Immediately truncates the batch, returns the trigger reason and evidence, hands control to the model, and activates recovery gating \\
\texttt{unavailable} & A required measurement is missing or invalid & Immediately truncates the batch, reports the missing key, and enforces Fine-step recovery until the measurement is restored \\
\bottomrule
\end{tabular}
\end{adjustbox}
\end{table*}

\begin{table*}[t]
\centering
\footnotesize
\renewcommand{\arraystretch}{1.08}
\setlength{\tabcolsep}{4pt}
\caption{Representative runtime monitoring and recovery cases from the main experiments.}
\label{tab:runtime_cases}
\begin{adjustbox}{max width=\linewidth}
\begin{tabular}{@{}L{0.17\textwidth}L{0.08\textwidth}L{0.31\textwidth}L{0.11\textwidth}L{0.27\textwidth}@{}}
\toprule
\textbf{Task (Layout)} & \textbf{Batches} & \textbf{Mechanism \& original expression} & \textbf{Req.~$\rightarrow$ Exec.} & \textbf{Truncation cause \& recovery} \\
\midrule
\textbf{Put Bottles Into Dustbin (L1)} & B17$\rightarrow$B18 & \textbf{Relative gripper displacement:} \texttt{abs(m('B.relative\_left.x')\allowbreak -ref('B.relative\_left.x')) > 0.025 or abs(m('B.relative\_left.z')\allowbreak -ref('B.relative\_left.z')) > 0.025} & 40 $\rightarrow$ 21 & Left-gripper $z$-displacement relative to \texttt{pink\_bottle} reached $\approx$27.35\,mm ($>$25\,mm); switched to Fine-step (8 steps). \\
\textbf{Build Tower (L3)} & B2$\rightarrow$B3 & \textbf{Pre-step distance lower bound:} \texttt{pred('lower\_surface\_\allowbreak distance:A:P') < 0.025} & 25 $\rightarrow$ 1 & Predicted lower bound from support block to left gripper pad was 24.607\,mm ($<$25\,mm); confirmed alarm and ran Fine-step (3 steps). \\
\textbf{Imitate Sorting Sequence (L2)} & B48$\rightarrow$B49 & \textbf{EEF position deviation:} \texttt{abs(m('R.x')-0.2365) > 0.008 or abs(m('R.y')+0.133) > 0.008 or m('R.z') < 0.800} & 25 $\rightarrow$ 5 & While lifting the watch, right-EEF $y$-deviation reached $\approx$8.49\,mm ($>$8\,mm); confirmed alarm and executed Fine-step lateral correction (3 steps). \\
\textbf{Classify Objects By Language (L2)} & B14$\rightarrow$B15 & \textbf{Placed-object surface excursion:} \texttt{m('O.z')-ref('O.z') > 0.015 or (m('O.x')-ref('O.x'))*\allowbreak(m('O.x')-ref('O.x'))+(m('O.y')\allowbreak -ref('O.y'))*\allowbreak(m('O.y')-ref('O.y')) > 0.0016} & 30 $\rightarrow$ 9 & Placed watch surface exceeded reference; visual review diagnosed tracking-drift \texttt{false\_positive}, re-registered inline, and resumed Large-step (25 req., 9 exec. to target). \\
\bottomrule
\end{tabular}
\end{adjustbox}
\end{table*}

\section{Perception and Low-Level Control Details}
\label{sec:perception_control}

This section supplements Sections~3.3--3.4 of the main paper with local perception configurations, multi-granularity control interfaces, and batch-level closed-loop feedback.

\subsection{Perception Configuration, Robot State, and Observation Boundary}

\begin{itemize}
  \item \textbf{Camera and depth inputs:} Three cameras (top-view \texttt{cam\_high}, left-wrist \texttt{cam\_left\_wrist}, and right-wrist \texttt{cam\_right\_wrist}) each output $640\times480$ \texttt{uint8} RGB images and \texttt{float32} optical-axis Z depth maps (in meters), paired with intrinsics $K$ and camera-to-environment extrinsics $T_{\text{env}\leftarrow\text{cam}}$. Depth maps are consumed exclusively by the local perception layer for 3D reconstruction; the decision model receives only raw RGB images.
  \item \textbf{SAM3 segmentation and 3D geometry:} Local perception segments and tracks objects using SAM3 (default sampling cap: 48 points). Back-projected 3D point clouds undergo 2-mm voxel downsampling, 1-cm cross-view depth-consistency verification, and statistical outlier filtering to extract surface geometry.
  \item \textbf{Robot state:} Proprioceptive state comes from simulator-provided 14-DoF joint and gripper-opening readings; end-effector poses (position and \texttt{wxyz} quaternion) are computed via forward kinematics in a unified control frame.
  \item \textbf{Observation boundary:} All RGB-D inputs, calibrations, and proprioceptive readings mirror standard real-robot sensors. Privileged simulator states---ground-truth object poses, segmentation masks, contact/collision states, and environment rewards---are strictly isolated from perception and decision-making. Benchmark scoring and termination remain independent of policy execution.
\end{itemize}

\subsection{Comparison of Large-step and Fine-step Control Interfaces}

Table~\ref{tab:control_modes} compares the two execution modes across action representation, step horizon, low-level tracking, and within-batch monitoring behavior. (Note: Figure~2 of the main paper illustrates end-effector orientation conceptually as Euler angles for readability; the actual runtime interface uses \texttt{wxyz} unit quaternions.)

\begin{table*}[t]
\centering
\footnotesize
\renewcommand{\arraystretch}{1.08}
\setlength{\tabcolsep}{5pt}
\caption{Comparison of Large-step and Fine-step control interfaces.}
\label{tab:control_modes}
\begin{adjustbox}{max width=\linewidth}
\begin{tabular}{@{}L{0.20\textwidth}L{0.37\textwidth}L{0.37\textwidth}@{}}
\toprule
\textbf{Dimension} & \textbf{Large-step (normal mode)} & \textbf{Fine-step (fine mode)} \\
\midrule
\textbf{Action representation} & World-frame EEF target pose (position and unit quaternion) and gripper opening $[0,1]$ & Local EEF relative-pose increments and gripper opening $[0,1]$ \\
\textbf{Step limit (Horizon)} & Constrained by remaining task/subtask budget (typically requests $\ge 25$ steps) & Short-step limit ($H_k \le 10$ steps), dynamically chosen from current visual evidence \\
\textbf{Low-level tracking} & Bounded IK solving with joint-space interpolation tracking & Local bounded EEF-increment tracking \\
\textbf{Within-batch monitoring} & Evaluated at every step; immediately truncates the batch upon \texttt{triggered} or \texttt{unavailable} & Logs step-level results and returns an aggregate at batch end without interrupting execution \\
\bottomrule
\end{tabular}
\end{adjustbox}
\end{table*}

\subsection{Batch Termination and Closed-Loop Feedback Process}

Corresponding to Algorithm~1 of the main paper, each completed batch returns the reached state, executed step count, and termination reason to drive the next decision:
\begin{itemize}
  \item \textbf{Normal termination and monitoring interruption:} A Large-step batch runs until target arrival, budget exhaustion, or a monitoring interruption. Based on the updated observation and monitoring evidence, the model either enters Fine-step for precise correction or revises the target and continues in Large-step.
  \item \textbf{Low-level motion limitation:} When a target pose violates workspace boundaries, joint limits, or kinematic constraints, the low-level controller safely terminates the batch. Partial progress returns the reached pose and solver diagnostics; completely unreachable targets return zero physical steps and a structured blocking diagnostic for replanning. Zero-step submissions still count as Action Batches while consuming no physical steps.
  \item \textbf{Interface-validation correction:} Occasional schema or parameter validation rejections return an error under the original observation state, prompting an in-place retry.
\end{itemize}

\section{RoboDojo Benchmark and Evaluation Protocol}
\label{sec:benchmark_protocol}

This section supplements Sections~4.1--4.3 of the main paper with benchmark task specifications, baseline result sources, and the ablation protocol.

\subsection{Evaluation Tasks, Layout Variations, and Native Scoring Rules}

We evaluate six bimanual tabletop manipulation tasks in RoboDojo covering all five official capability dimensions (\emph{Generalization}, \emph{Memory}, \emph{Precision}, \emph{Long-Horizon}, and \emph{Open}). All methods are evaluated across \textbf{five layouts per task under Seed~0}: the seed selects the candidate scene pool, and each layout-reset seed uniquely determines initial object categories, poses, and physical properties (used strictly for reproducible initialization, never as policy input). Native environment scores follow a stage-transition rubric (0--100, normalized to $[0,1]$ in Table~1 of the main paper) independent of the model's self-assessment (Table~\ref{tab:robodojo_tasks}).

\begin{table*}[t]
\centering
\footnotesize
\renewcommand{\arraystretch}{1.08}
\setlength{\tabcolsep}{4pt}
\caption{RoboDojo evaluation tasks, layout variations, step budgets, and native stage-scoring rubrics.}
\label{tab:robodojo_tasks}
\begin{adjustbox}{max width=\linewidth}
\begin{tabular}{@{}L{0.18\textwidth}L{0.23\textwidth}L{0.20\textwidth}L{0.08\textwidth}L{0.25\textwidth}@{}}
\toprule
\textbf{Task name} & \textbf{Core objective} & \textbf{Layout variations} & \textbf{Max steps} & \textbf{Native stage scores (0--100)} \\
\midrule
\textbf{Classify Objects By Language} & Place three object categories into left, middle, and right baskets per language instruction, then reset arms & Category combinations, instance counts, initial poses & 1,100 & 10 / 40 / 100 (1--3 categories placed with grippers open) \\
\textbf{Build Tower} & Stack wooden blocks and boards into a multi-level tower layer by layer & Initial block and board positions/orientations & 1,050 & 10 / 30 / 100 (layer-by-layer stability criteria met) \\
\textbf{Imitate Sorting Sequence} & Observe and memorize the demonstrated placement order of five objects, then replicate it & Demonstration order, initial tabletop arrangement & 1,600 & 5 / 15 / 30 / 50 / 100 (1--5 objects placed in order, grippers open) \\
\textbf{Arrange Largest Number} & Arrange 4 or 5 digit blocks left-to-right on the mat to form the largest integer & Digit values, block count (4/5), standard/random layout & 1,050 & 4 blocks: 5 / 15 / 30 / 100; 5 blocks: 5 / 15 / 25 / 40 / 100 \\
\textbf{Put Bottles Into Dustbin} & Pick up four bottles and drop them into the dustbin, using bimanual handover if needed & Bottle styles, tabletop spatial distribution & 700 & 10 / 25 / 40 / 100 (1--4 bottles placed, grippers open) \\
\textbf{Pack Objects Into Box} & Pack four objects into an upright cardboard box with all front faces pointing left & Object combinations, poses, standard/random variants & 1,300 & 10 / 25 / 50 / 100 (box upright and 1--4 objects packed, grippers open) \\
\bottomrule
\end{tabular}
\end{adjustbox}
\end{table*}

\subsection{Baseline Result Sources and System-Level Comparison}

\paragraph{Evaluation Setting and Sources.}
All four methods in Table~1 of the main paper share the \textbf{Seed~0, five-layout-per-task} protocol:
\begin{itemize}
  \item \textbf{GAP Direct and RoboDawn:} Official evaluation results under this setting (normalized from 0--100 to $[0,1]$). Following official reporting, RoboDawn replaces Layout~4 of \emph{Imitate Sorting Sequence} with a stable layout.
  \item \textbf{PhysicalRSI:} External reference evaluated privately under the exact same Seed~0, five-layout protocol.
\end{itemize}

\paragraph{Configuration Differences and Comparison Scope.}
Table~\ref{tab:baseline_configs} contrasts each method's reasoning strength, action representation, and observation/tool permissions. Tables~1--2 of the main paper are positioned as system-level comparisons, testing whether MATCH---operating at lower reasoning strength (\texttt{Low}), zero-shot, and without predefined skill libraries---outperforms a stronger-reasoning (\texttt{XHigh}) direct-control baseline and competes with tool- or demonstration-augmented systems. Controlled component-level attribution is isolated in Table~3.

\begin{table*}[t]
\centering
\footnotesize
\renewcommand{\arraystretch}{1.08}
\setlength{\tabcolsep}{5pt}
\caption{System-level configuration comparison across evaluated methods.}
\label{tab:baseline_configs}
\begin{adjustbox}{max width=\linewidth}
\begin{tabular}{@{}L{0.15\textwidth}L{0.16\textwidth}L{0.29\textwidth}L{0.34\textwidth}@{}}
\toprule
\textbf{Method} & \textbf{Reasoning \& demos} & \textbf{Action representation \& control} & \textbf{Observation \& tool permissions} \\
\midrule
\textbf{MATCH (Ours)} & \texttt{Low}, Zero-Shot & Variable-step, multi-granularity EEF control & Three RGB views + local depth/SAM3 perception and step-level conditional monitoring (no skill library) \\
\textbf{GAP Direct} & \texttt{XHigh}, Zero-Shot & Fixed 5-step, 14-DoF direct EEF control without hierarchical planning or monitoring & Three RGB views for direct decisions; no VLA or local monitoring \\
\textbf{PhysicalRSI} & \texttt{XHigh}, Zero-Shot & VLA combined with discrete operation tools & Predefined discrete operation-skill interfaces \\
\textbf{RoboDawn} & \texttt{High}, 1-Demo & Discrete operation tools guided by a single expert demonstration & Discrete operation tools and task-demonstration input \\
\bottomrule
\end{tabular}
\end{adjustbox}
\end{table*}

\subsection{Table~3 Ablation Protocol}

All ablation variants in Table~3 were executed independently in the same environment using \textbf{GPT-6 Astra (Low)}, controlling for reasoning strength to isolate each component's incremental contribution. We select two complementary tasks: \emph{Put Bottles Into Dustbin} (long-range transport with slip-prone grasps) and \emph{Build Tower} (high-precision contact alignment with cross-layer error accumulation). Complete variant configurations and mechanism boundaries are detailed in Appendix~\ref{sec:ablations} (Table~\ref{tab:ablation_matrix}).

\section{Per-Layout Results}
\label{sec:per_layout}

Table~\ref{tab:per_layout_results} reports per-layout task scores, action batches, and physical control steps across all six tasks and five layouts under Seed~0.

\begin{table*}[tp]
\centering
\scriptsize
\renewcommand{\arraystretch}{1.05}
\setlength{\tabcolsep}{6pt}
\caption{Per-layout task scores, action batches, and control steps across all six RoboDojo tasks (Seed~0).}
\label{tab:per_layout_results}
\begin{adjustbox}{max width=\linewidth}
\begin{tabular}{@{}llrrrrrr@{}}
\toprule
& & \multicolumn{3}{c}{\textbf{MATCH (Ours)}} & \multicolumn{3}{c}{\textbf{GAP Direct}} \\
\cmidrule(lr){3-5} \cmidrule(lr){6-8}
\textbf{Task} & \textbf{Layout} & \textbf{Score} & \textbf{Batches} & \textbf{Steps} & \textbf{Score} & \textbf{Batches} & \textbf{Steps} \\
\midrule
\textbf{Classify Objects By Language} & L0 & 1.00 & 145 & 1,080 & 0.40 & 220 & 1,100 \\
& L1 & 0.40 & 173 & 1,097 & 0.00 & 56 & 257 \\
& L2 & 1.00 & 140 & 995 & 1.00 & 159 & 791 \\
& L3 & 0.00 & 172 & 1,059 & 0.00 & 283 & 1,100 \\
& L4 & 1.00 & 145 & 982 & 1.00 & 196 & 978 \\
& \textbf{Mean} & \textbf{0.68} & \textbf{155.0} & \textbf{1,042.6} & \textbf{0.48} & \textbf{182.8} & \textbf{845.2} \\
\midrule
\textbf{Build Tower} & L0 & 1.00 & 148 & 992 & 0.10 & 210 & 1,050 \\
& L1 & 0.10 & 138 & 1,050 & 0.00 & 210 & 1,050 \\
& L2 & 0.10 & 146 & 1,050 & 0.30 & 210 & 1,050 \\
& L3 & 0.30 & 156 & 1,050 & 0.10 & 211 & 1,050 \\
& L4 & 0.30 & 142 & 1,050 & 0.10 & 210 & 1,050 \\
& \textbf{Mean} & \textbf{0.36} & \textbf{146.0} & \textbf{1,038.4} & \textbf{0.12} & \textbf{210.2} & \textbf{1,050.0} \\
\midrule
\textbf{Imitate Sorting Sequence} & L0 & 1.00 & 149 & 1,425 & 0.00 & 90 & 450 \\
& L1 & 1.00 & 155 & 1,510 & 0.00 & 22 & 104 \\
& L2 & 0.30 & 169 & 1,600 & 0.00 & 76 & 378 \\
& L3 & 1.00 & 89 & 1,070 & 0.00 & 11 & 54 \\
& L4 & ---$^{\ddagger}$ & 142 & ---$^{\dagger}$ & 0.00 & 21 & 98 \\
& \textbf{Mean} & \textbf{0.825}$^{\ddagger}$ & \textbf{140.8} & \textbf{1,401.3}$^{\dagger}$ & \textbf{0.00} & \textbf{44.0} & \textbf{216.8} \\
\midrule
\textbf{Arrange Largest Number} & Std L0 & 1.00 & 83 & 700 & 1.00 & 176 & 877 \\
& Std L1 & 1.00 & 67 & 433 & 1.00 & 147 & 735 \\
& Rand L0 & 1.00 & 66 & 578 & 0.30 & 210 & 1,050 \\
& Rand L1 & 1.00 & 79 & 515 & 0.25 & 210 & 1,050 \\
& Rand L2 & 1.00 & 65 & 550 & 0.30 & 210 & 1,050 \\
& \textbf{Mean} & \textbf{1.00} & \textbf{72.0} & \textbf{555.2} & \textbf{0.57} & \textbf{190.6} & \textbf{952.4} \\
\midrule
\textbf{Put Bottles Into Dustbin} & L0 & 1.00 & 74 & 686 & 0.40 & 140 & 700 \\
& L1 & 1.00 & 45 & 457 & 1.00 & 121 & 602 \\
& L2 & 0.40 & 78 & 678 & 0.00 & 140 & 700 \\
& L3 & 1.00 & 67 & 587 & 0.00 & 140 & 700 \\
& L4 & 0.25 & 88 & 691 & 0.40 & 140 & 700 \\
& \textbf{Mean} & \textbf{0.73} & \textbf{70.4} & \textbf{619.8} & \textbf{0.36} & \textbf{136.2} & \textbf{680.4} \\
\midrule
\textbf{Pack Objects Into Box} & L0 & 0.50 & 159 & 1,300 & 0.50 & 260 & 1,300 \\
& L1 & 1.00 & 179 & 1,300 & 1.00 & 222 & 1,109 \\
& L2 & 0.50 & 172 & 1,300 & 0.50 & 260 & 1,300 \\
& L3 & 0.25 & 172 & 1,300 & 0.00 & 260 & 1,300 \\
& L4 & 0.25 & 177 & 1,300 & ---$^{\ddagger}$ & 192 & 960 \\
& \textbf{Mean} & \textbf{0.50} & \textbf{171.8} & \textbf{1,300.0} & \textbf{0.50}$^{\ddagger}$ & \textbf{238.8} & \textbf{1,193.8} \\
\bottomrule
\end{tabular}
\end{adjustbox}
\vspace{3pt}
\par\noindent\begin{minipage}{\textwidth}
\scriptsize
$^{\ddagger}$ \textbf{Missing scores (logging anomaly):} Due to logging anomalies in the official evaluation script, the raw evaluation records for MATCH L4 (\emph{Imitate Sorting Sequence}) and GAP Direct L4 (\emph{Pack Objects Into Box}) did not emit valid final scores (marked \texttt{---}), even though both episodes executed normally (142 and 192 action batches, respectively, rather than crashing prematurely). Task means (\texttt{0.825}, rounded to \texttt{0.83} in Table~1 of the main paper, and \texttt{0.50}) are therefore averaged over the four valid layouts L0--L3, whereas batch counts average all five layouts.\\
$^{\dagger}$ \textbf{Missing steps:} For the same official script logging anomaly, the raw log for MATCH L4 (\emph{Imitate Sorting Sequence}) lacked a valid step count; the mean (\texttt{1,401.3}) averages L0--L3.
\end{minipage}
\end{table*}

\paragraph{Why GAP Direct Exhibits Low Batch Counts on \emph{Imitate Sorting Sequence}.}
Consistent with Section~4.2 of the main paper, per-layout step counts confirm that GAP Direct's lower batch count (44.0 vs.\ 140.8) stems from early failure rather than higher decision efficiency: GAP Direct advances only 450 / 104 / 378 / 54 / 98 steps across the five layouts (mean 216.8 steps, $<$13.6\% of the 1,600-step budget) and scores 0.00 on every run. By contrast, MATCH's step-level monitoring corrects intermediate deviations in time, sustaining 1,070--1,600 steps and achieving strong task completion.

\section{Full Ablations and Sensitivity Analysis}
\label{sec:ablations}

This section supplements Section~4.3 and Table~3 of the main paper with the complete ablation configuration matrix, mechanism boundaries, and key parameter observations.

\subsection{Complete Configuration Matrix for Ablations and Controls}
\label{sec:ablation_matrix}

Table~\ref{tab:ablation_matrix} consolidates the planning, monitoring, prompt, and interface differences across all Table~3 ablation variants and the main-experiment direct-control baseline.

\begin{table*}[t]
\centering
\footnotesize
\renewcommand{\arraystretch}{1.08}
\setlength{\tabcolsep}{4pt}
\caption{Complete configuration matrix for all ablation variants and direct-control baselines.}
\label{tab:ablation_matrix}
\begin{adjustbox}{max width=\linewidth}
\begin{tabular}{@{}L{0.18\textwidth}L{0.21\textwidth}L{0.22\textwidth}L{0.21\textwidth}L{0.12\textwidth}@{}}
\toprule
\textbf{Variant} & \textbf{Planning / context} & \textbf{Monitoring / perception} & \textbf{Prompt \& interface} & \textbf{Reasoning} \\
\midrule
\textbf{MATCH (Full)} & Full hierarchy (\texttt{Mission} / \texttt{Subtask} / \texttt{Operate}) and switching protocol & SAM3 perception + step-level conditional monitoring & Full planning, multi-granularity action, and monitoring interfaces & GPT-6 Astra (\texttt{Low}) \\
\textbf{MATCH (w/o Planning)} & Overall goal (\texttt{mission\_outline}) only; removes \texttt{Subtask}/\texttt{Operate} & SAM3 perception + step-level conditional monitoring & Overall-task prompt only; removes planning fields and ID-binding checks & GPT-6 Astra (\texttt{Low}) \\
\textbf{MATCH (w/o Monitoring)} & Full hierarchical planning and operation specifications & Automatic monitoring pipeline and interruption disabled & Removes monitoring instructions, registration, review, and protection fields & GPT-6 Astra (\texttt{Low}) \\
\textbf{GAP (Fixed Step = 10)} & Direct control + work history & No automatic runtime monitoring & Reuses GAP Direct prompt with fixed horizon increased from 5 to 10 steps & GPT-6 Astra (\texttt{Low}) \\
\textbf{GAP Direct (Fixed Step = 5)} & Direct control + work history & Three RGB views + robot state; no runtime monitoring & Fixed 5-step direct-control prompt and interface & GPT-6 Astra (\texttt{XHigh}) \\
\bottomrule
\end{tabular}
\end{adjustbox}
\end{table*}

\subsection{Mechanism Boundaries of Core Ablation Variants}

\begin{itemize}
  \item \textbf{GAP (Fixed Step = 10):} Under unified \texttt{Low} reasoning, statically extends direct control from 5 to 10 steps without hierarchical planning, adaptive horizons, or within-batch interruption. Because it cannot shorten steps or brake midway during contact alignment, it suffers from open-loop blind spots during fine manipulation, causing sharp score degradation.
  \item \textbf{MATCH (w/o Planning):} Retains all physical observations (three RGB views, robot state, SAM3 perception, and monitoring feedback) while removing structured planning (\texttt{Subtask} decomposition, \texttt{Operate} specifications, switching/reset protocols, and subtask budgets). Without persistent local constraints across batches, the model repeatedly re-infers strategy each turn, increasing decision batches and hurting long-horizon scores.
  \item \textbf{MATCH (w/o Monitoring):} Retains hierarchical planning and Large-/Fine-step action generation while disabling automated runtime monitoring (within-batch interruption, pre-step kinematic prediction, monitoring reports, and recovery gating), isolating the value of step-level closed-loop intervention over batch-level planning alone.
\end{itemize}

\subsection{Key-Parameter Sensitivity and Runtime Observations}

Main-experiment and exploratory logs reveal three key parameter behaviors:
\begin{itemize}
  \item \textbf{Fine-step limit:} Defaulting to 10 steps (observed maximum: 10 steps/batch), enabling rapid visual feedback during contact alignment and anomaly recovery.
  \item \textbf{Large-step horizon:} Advances adaptively with lower bound $\min(25, \text{remaining steps})$ (observed maximum: 44 steps/batch). In an exploratory run capping both modes at $\min(10, \text{remaining steps})$ while retaining full planning and monitoring, long-range free-space motion fragments into short batches and frequently triggers partial-progress fallback, substantially inflating decision overhead.
  \item \textbf{Adaptive monitoring thresholds:} The model autonomously calibrates thresholds to contact risk, spanning millimeter to centimeter scales---e.g., 25\,mm relative slip for bottle insertion, 25\,mm pre-step proximity for tower building, 10\,mm support disturbance for upper tower layers, and 35\,mm proximity for digit blocks---demonstrating phase-aware tolerance adaptation.
\end{itemize}

\subsection{Attribution Summary}

Under unified \texttt{Low} reasoning, Table~3 cleanly isolates each component's contribution: hierarchical planning (\texttt{w/o Planning} $\rightarrow$ \texttt{Full}) reduces redundant decisions and boosts long-horizon completion; step-level runtime monitoring (\texttt{w/o Monitoring} $\rightarrow$ \texttt{Full}) intercepts free-space slippage and multi-layer contact drift; and \texttt{GAP (Fixed Step = 10)} confirms that static step lengthening cannot trade off efficiency and precision. Crucially, \texttt{MATCH (w/o Planning)} retains the exact same RGB-D/SAM3 perception and monitoring pipeline as \texttt{MATCH (Full)} yet drops sharply in performance (mean score 0.270 vs.\ 0.545), directly ruling out the hypothesis that MATCH's gains stem solely from 3D perception. Because GAP Direct and MATCH in Table~1 differ in reasoning strength and system configuration, Table~1 reflects end-to-end system gains, whereas component-level attribution rests on Table~3.

\section{Runtime, Latency, and Token Analysis}
\label{sec:runtime_analysis}

\begin{figure*}[t!]
    \centering
    \includegraphics[width=0.95\textwidth]{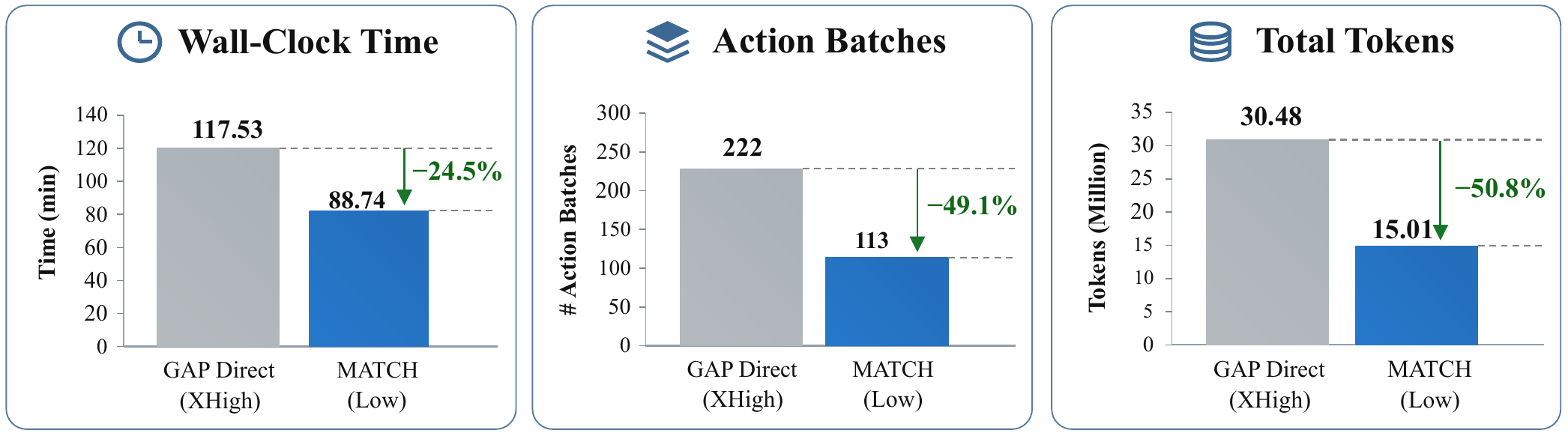}
    \caption{Episode-level comparison of wall-clock time, action batches, and total token consumption between GAP Direct (\texttt{XHigh}) and MATCH (\texttt{Low}) on an additional instrumented profiling run of \emph{Pack Objects Into Box} (L1), where both methods achieve a task score of 1.00.}
    \label{fig:runtime_comparison}
\end{figure*}

Action Batches quantify the frequency of high-level decisions. Although this metric does not correspond directly to wall-clock time, it is unaffected by fluctuations in network latency and server load, making it a central measure of multi-timescale scheduling efficiency. Because the main-experiment logs in Table~\ref{tab:per_layout_results} did not record wall-clock timing or token telemetry, we conduct an additional instrumented single-episode profiling run on \emph{Pack Objects Into Box} (L1) to measure end-to-end runtime, API token usage, and per-invocation component latencies. These measurements illustrate how reducing action decision batches translates into practical runtime and context-efficiency gains.

\subsection{End-to-End Runtime and Component-Level Latency}

In the instrumented single-episode comparison on \emph{Pack Objects Into Box} (L1) (Figure~\ref{fig:runtime_comparison}), both methods achieve a task score of 1.00, while MATCH (\texttt{Low} reasoning effort) completes execution in \textbf{88.74 minutes} compared with \textbf{117.53 minutes} for GAP Direct (\texttt{XHigh} reasoning effort), saving \textbf{28.79 minutes (24.5\%)}. Over this profiled run, MATCH requires \textbf{113 action batches} compared with \textbf{222} for GAP Direct (a reduction of \textbf{49.1\%}); note that due to stochastic model sampling, this instrumented rerun uses fewer batches than the main-experiment run on the same layout in Table~\ref{tab:per_layout_results} (179 batches, also scoring 1.00), so we further analyze per-invocation latencies below to illustrate why reduced decision frequency translates into general runtime savings.

\begin{figure}[!htb]
    \centering
    \includegraphics[width=\linewidth]{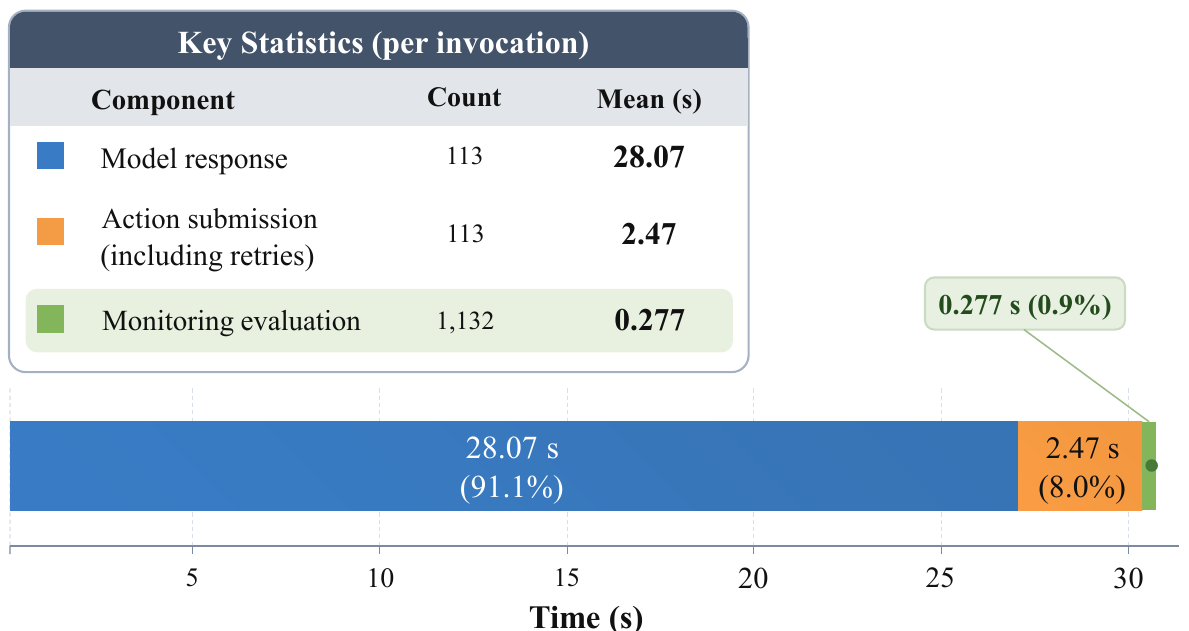}
    \caption{Average per-invocation latency of model responses, action submissions, and monitoring evaluations in MATCH. Percentage shares are normalized by the sum of the three component means rather than cumulative episode runtime.}
    \label{fig:latency_breakdown}
\end{figure}

Component-level latency measurements from the same profiled run (Figure~\ref{fig:latency_breakdown}) further show that model responses are substantially more time-consuming per invocation than local processing:

\begin{itemize}
  \item \textbf{Model response:} 28.07\,s per response on average.
  \item \textbf{Action submission (local tool processing, including retries):} 2.47\,s per invocation on average, including mean latencies of 1.71\,s for inline object registration and 0.473\,s for retained IK path-preparation reports.
  \item \textbf{Monitoring evaluation:} 0.277\,s per evaluation on average.
\end{itemize}

The mean latency of a model response (28.07\,s) is roughly an order of magnitude greater than that of an action submission (2.47\,s) and two orders of magnitude greater than an individual monitoring evaluation (0.277\,s). By combining adaptive Large-step execution with task-conditioned monitoring, MATCH reduces costly high-level action decisions by 24.6\% on average across the main experiments. Even when monitoring triggers additional corrective batches in contact-intensive tasks, individual local evaluations remain lightweight relative to model responses.

\subsection{Episode-Level Token Consumption}

\begin{figure}[t!]
    \centering
    \includegraphics[width=0.86\linewidth]{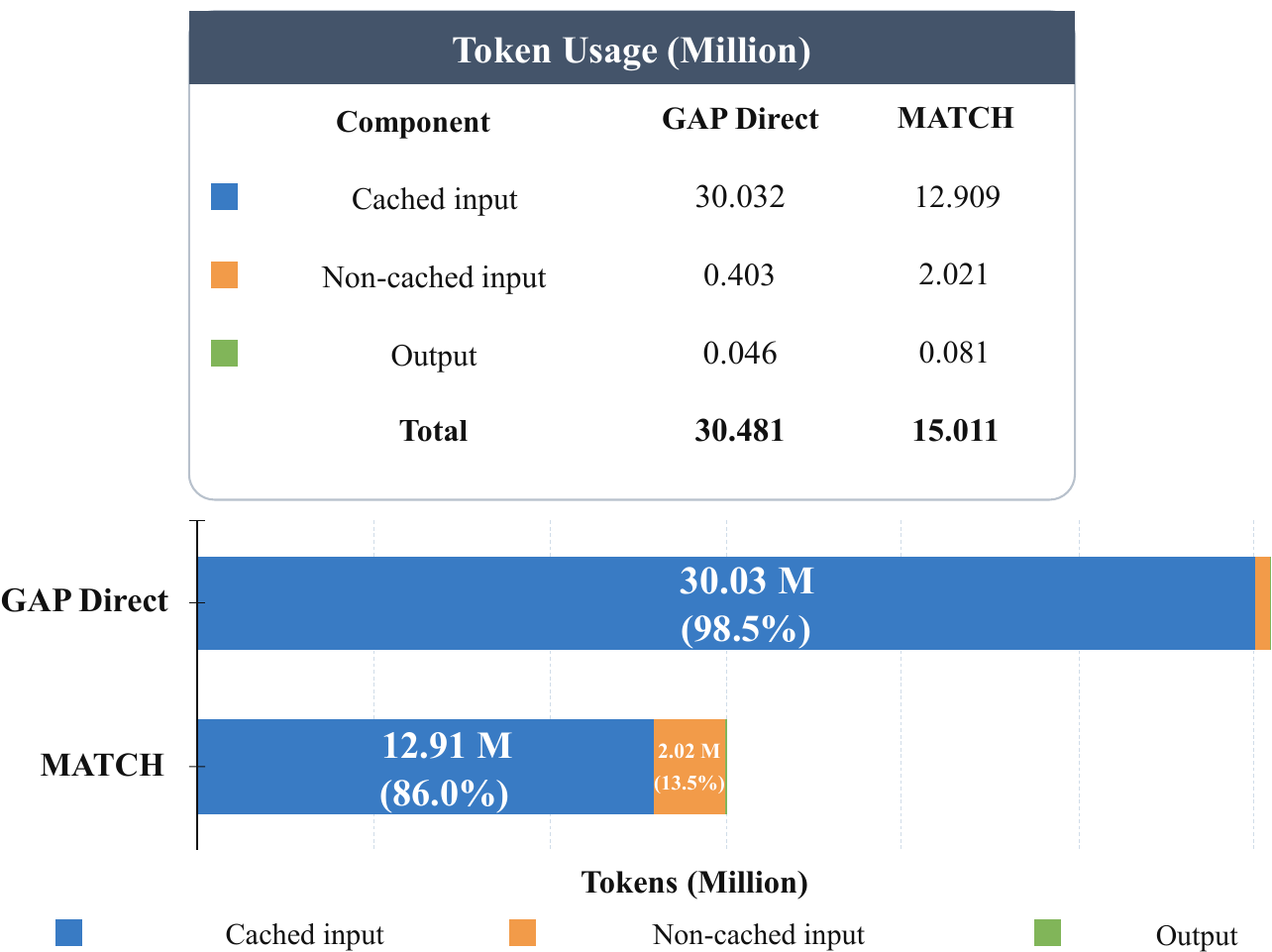}
    \vspace{-4pt}
    \caption{Token consumption breakdown for GAP Direct (\texttt{XHigh}) and MATCH (\texttt{Low}) on the instrumented \emph{Pack Objects Into Box} (L1) run (cached input is a subset of input; reasoning tokens are included in output).}
    \label{fig:token_breakdown}
\end{figure}

For the same profiled episode (Figure~\ref{fig:token_breakdown}), MATCH (\texttt{Low} reasoning effort) substantially reduces both input and total token consumption relative to the original GAP Direct evaluation (\texttt{XHigh} reasoning effort):

\begin{itemize}
  \item \textbf{Input tokens:} 14,929,823 for MATCH versus 30,435,661 for GAP Direct, a reduction of \textbf{50.9\%}.
  \item \textbf{Cached input tokens:} 12,908,800 for MATCH versus 30,032,384 for GAP Direct.
  \item \textbf{Non-cached input tokens:} 2,021,023 for MATCH versus 403,277 for GAP Direct.
  \item \textbf{Output tokens:} 81,274 for MATCH versus 45,543 for GAP Direct.
  \item \textbf{Total tokens:} 15,011,097 for MATCH versus 30,481,204 for GAP Direct, a reduction of \textbf{50.8\%}.
\end{itemize}

This reduction aligns with fewer action decisions and subtask-scoped session resets, which keep historical context compact and prevent excessive token growth within a single long-running session. Reinitializing context at subtask boundaries increases non-cached input tokens, while jointly generating planning updates, actions, and monitoring specifications increases output tokens; together, this design maintains a compact context and cuts total token consumption by more than half while preserving the task score.

Across the main experiments, MATCH also achieves a higher mean task score than GAP Direct, demonstrating strong overall efficiency across task performance, execution time, and token usage.

\section{Qualitative Traces and Failure Analysis}
\label{sec:qualitative_failure}

This section analyzes representative intervention traces, interruption statistics, and failure modes across our RoboDojo rollouts using logged execution telemetry and camera views. Throughout, \texttt{B} denotes submitted action batches and \texttt{t} denotes executed environment control steps.

\subsection{Representative Mid-Batch Intervention Traces}

\begin{figure*}[p]
    \centering
    \includegraphics[width=0.92\textwidth, height=0.41\textheight, keepaspectratio]{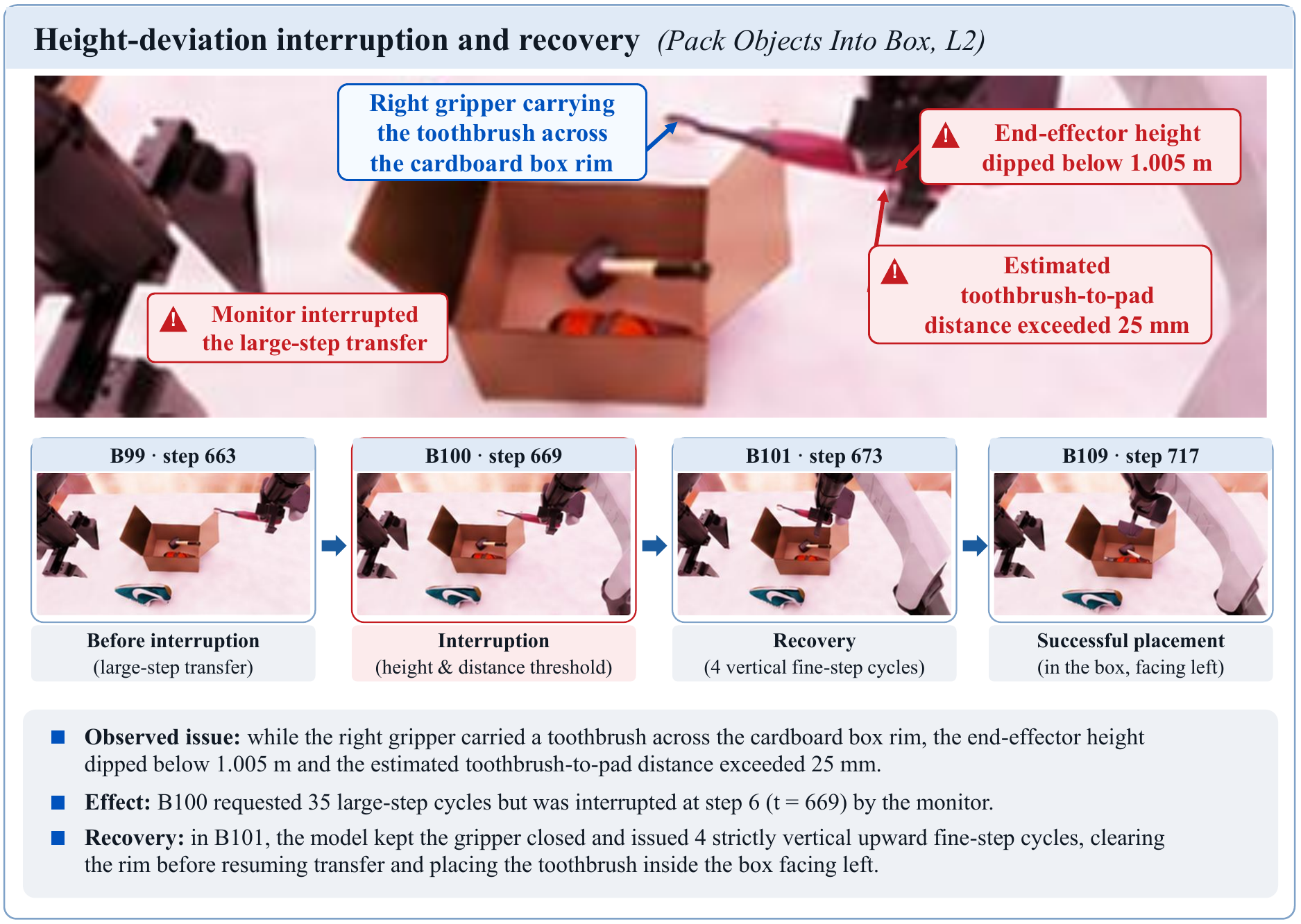}
    \caption{Height-deviation interruption and recovery during toothbrush placement in \emph{Pack Objects Into Box} (L2).}
    \label{fig:height_recovery}
    \vspace{12pt}
    \includegraphics[width=0.92\textwidth, height=0.41\textheight, keepaspectratio]{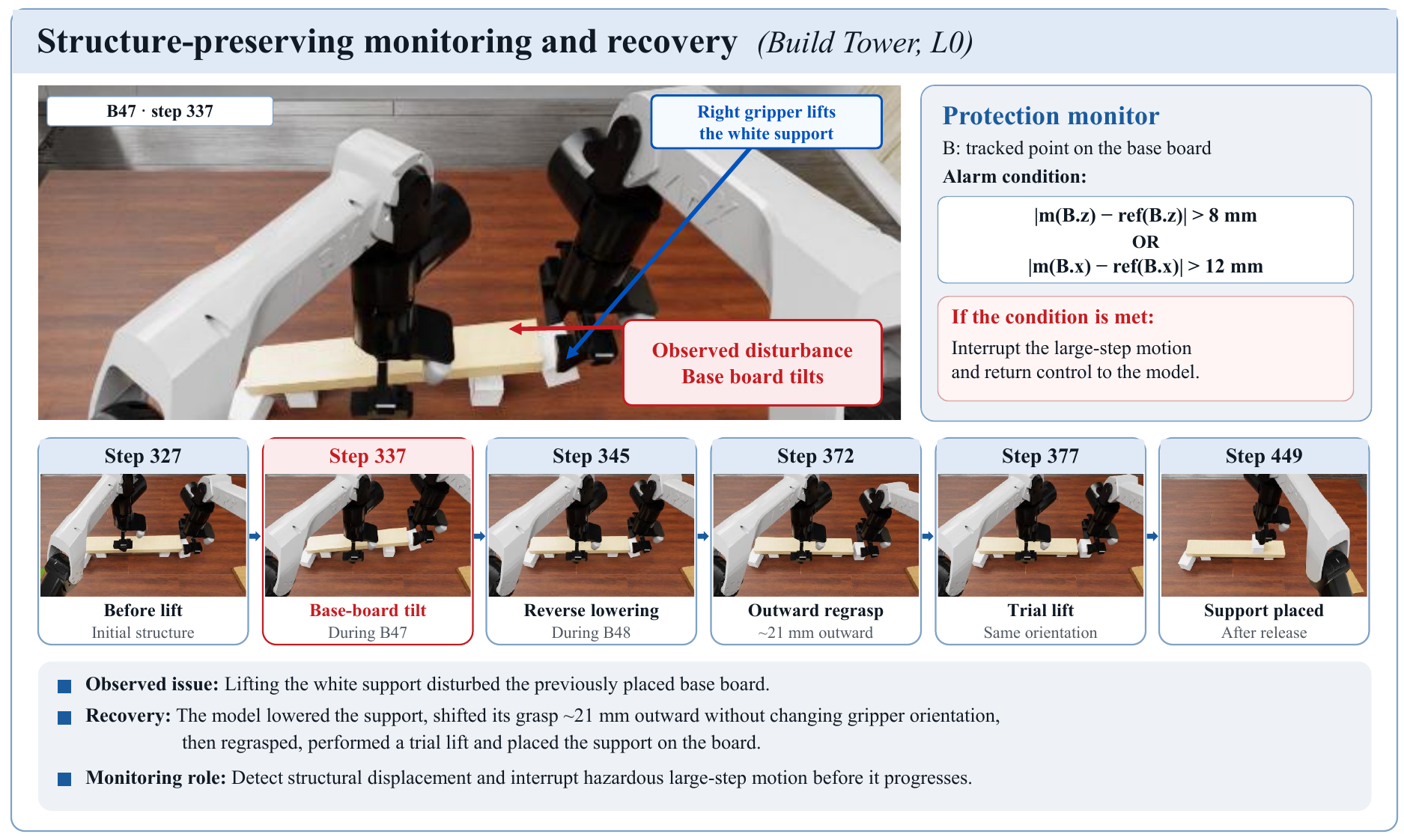}
    \caption{Structure-preserving monitoring and recovery during support-block manipulation in \emph{Build Tower} (L0).}
    \label{fig:tower_recovery}
\end{figure*}

\paragraph{Height-Deviation Interruption and Recovery (\emph{Pack Objects Into Box}, L2).}
As illustrated in Figure~\ref{fig:height_recovery}, while the right gripper carried a toothbrush across the cardboard box rim, \texttt{B100} requested 35 Large-step cycles but was interrupted at step~6 (\texttt{t = 669}): the end-effector height dipped below 1.005\,m and the estimated toothbrush-to-pad distance exceeded 25\,mm. In \texttt{B101}, the model kept the gripper closed and issued 4 strictly vertical upward Fine-step cycles, clearing the rim before resuming transfer and placing the toothbrush inside the box facing left.

\paragraph{Structure-Preserving Monitoring and Recovery (\emph{Build Tower}, L0).}
As illustrated in Figure~\ref{fig:tower_recovery}, during the lifting of a white support block in \texttt{B47}, one end of the previously assembled base board was unintentionally lifted, threatening the stability of the existing structure. The protection monitor tracks a point $B$ on the base board, with an alarm condition defined as a vertical displacement exceeding 8\,mm or a lateral ($x$-axis) displacement exceeding 12\,mm. In \texttt{B48}, the model reversed the lifting motion, restoring the board to a near-horizontal configuration. It subsequently shifted the grasping position approximately 21\,mm outward while maintaining the gripper orientation, regrasped the support, performed a trial lift, and successfully placed it on the base board. This case illustrates how structure-aware monitoring and adaptive regrasping can mitigate unintended disturbances to previously completed subtasks.

\subsection{Per-Layout Breakdown of Mid-Batch Interruptions}

To substantiate the multi-granularity scheduling statistics in Section~4.3 of the main paper, Table~\ref{tab:interruption_audit} audits all five \emph{Put Bottles Into Dustbin} (L0--L4) main-experiment rollouts. An event is counted as an interruption when a Large-step batch executes at least one physical step and ends early due to a monitor trigger before exhausting its requested limit or executable plan (excluding zero-step blocks).

\begin{table}[!htb]
\centering
\footnotesize
\renewcommand{\arraystretch}{1.08}
\setlength{\tabcolsep}{3pt}
\caption{Per-layout audit of monitor-related mid-batch interruptions across \emph{Put Bottles Into Dustbin} (L0--L4, MATCH Full).}
\label{tab:interruption_audit}
\begin{adjustbox}{max width=\linewidth}
\begin{tabular}{@{}lrrrr@{}}
\toprule
\textbf{Layout} & \textbf{Pos.-step Large} & \textbf{Interruptions} & \textbf{Next: Fine} & \textbf{Next: Large} \\
\midrule
L0 & 36 & 18 & 6 & 12 \\
L1 & 22 & 4 & 4 & 0 \\
L2 & 32 & 7 & 6 & 1 \\
L3 & 33 & 5 & 5 & 0 \\
L4 & 32 & 7 & 7 & 0 \\
\midrule
\textbf{Total} & \textbf{155} & \textbf{41 (26.5\%)} & \textbf{28 (68.3\%)} & \textbf{13 (31.7\%)} \\
\bottomrule
\end{tabular}
\end{adjustbox}
\end{table}

Across the 41 interruptions, 39 were triggered by abnormality predicates (including pre-step predictive bounds) and 2 by unavailable measurements. All 28 immediate Fine-step follow-ups executed to their requested short horizons. Among the 13 immediate Large-step follow-ups, 5 reached the target pose, 2 executed partial routes, and 6 were interrupted again.

\subsection{Limitations and Failure Attribution}

Table~\ref{tab:failure_attribution} summarizes the four recurring monitoring- and perception-level limitations observed in our execution logs.

\begin{table*}[t!]
\centering
\scriptsize
\renewcommand{\arraystretch}{1.05}
\setlength{\tabcolsep}{5pt}
\caption{Taxonomy of recurring system limitations and failure sources supported by logged execution traces.}
\label{tab:failure_attribution}
\begin{adjustbox}{max width=\linewidth}
\begin{tabular}{@{}L{0.20\textwidth}L{0.76\textwidth}@{}}
\toprule
\textbf{Failure / limitation source} & \textbf{Representative logged trace and mechanism} \\
\midrule
\textbf{Visual tracking drift} & \emph{Put Bottles Into Dustbin} (L0, \texttt{B27}--\texttt{B28}): During the receiving arm's approach, \texttt{B27} was interrupted at 2/25 steps (\texttt{t = 298}). Visual review showed the tracker had switched identity to a nearby yellow bottle while the held white bottle remained stable; \texttt{B28} dismissed the affected predicate and resumed the approach. \\
\textbf{Measurement unavailability} & \emph{Put Bottles Into Dustbin} (L3, \texttt{B27} and \texttt{B61}): Both batches halted early (\texttt{t = 248}, \texttt{550}) because the required bottle-to-bin distance became unavailable, triggering conservative Fine-step fallback. A stationary refresh restored tracking after \texttt{B27}, whereas \texttt{B61} remained unavailable after the first Fine-step recovery batch. \\
\textbf{Monitor coverage gap} & \emph{Pack Objects Into Box} (L0, \texttt{B31}): The cardboard box visibly shifted during empty-hand departure, yet \texttt{B31} completed normally (33/45 steps, \texttt{t = 359}) because the active rules monitored finger--box proximity and vertical shoe motion rather than direct box displacement. The subsequent turn detected the shift visually and scheduled a re-check. \\
\textbf{Physical retention loss} & \emph{Pack Objects Into Box} (L0, \texttt{B148}): Truncated at 2/25 steps when the toy car's tracked hand-relative displacement exceeded 15\,mm. Wrist-camera review confirmed grasp loss; a subsequent regrasp briefly supported a trial lift before losing the car again, leaving placement incomplete. \\
\bottomrule
\end{tabular}
\end{adjustbox}
\end{table*}

\begin{figure*}[b!]
    \centering
    \includegraphics[width=0.92\textwidth, height=0.58\textheight, keepaspectratio]{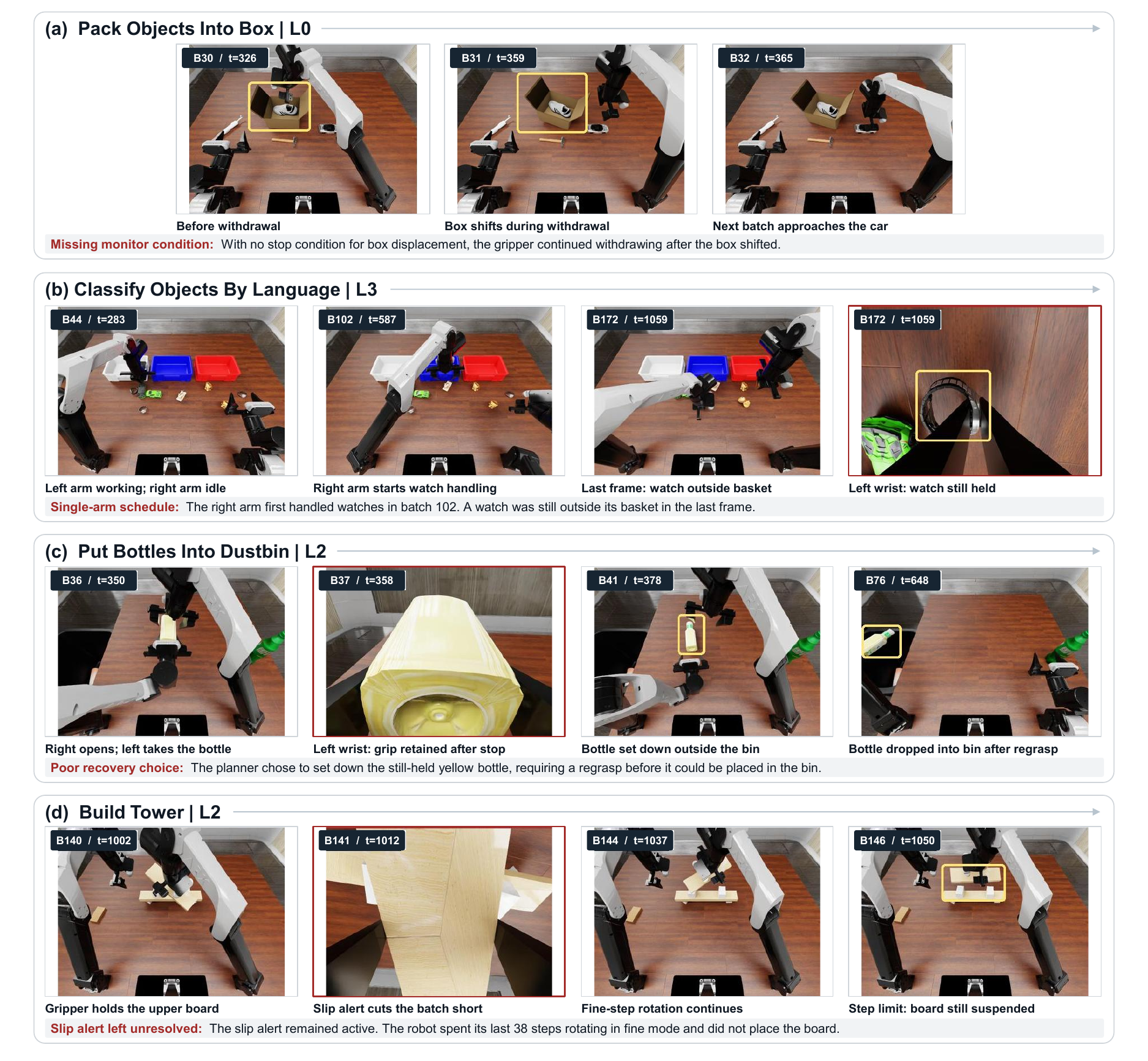}
    \caption{Representative failure cases across four RoboDojo tasks: (a) missing monitor condition for box displacement in \emph{Pack Objects Into Box} (L0); (b) delayed bimanual scheduling in \emph{Classify Objects By Language} (L3); (c) suboptimal post-interruption recovery choice in \emph{Put Bottles Into Dustbin} (L2); and (d) unresolved persistent slip alert near the step limit in \emph{Build Tower} (L2).}
    \label{fig:failure_cases}
\end{figure*}

Figure~\ref{fig:failure_cases} complements Table~\ref{tab:failure_attribution} with four representative end-to-end failure trajectories spanning both monitoring gaps and higher-level planning or recovery bottlenecks: (a) a missing box-displacement monitor condition during arm withdrawal in \emph{Pack Objects Into Box} (L0, corresponding to the monitor coverage gap in Table~\ref{tab:failure_attribution}); (b) delayed bimanual utilization in \emph{Classify Objects By Language} (L3), where the right arm did not begin handling watches until \texttt{B102}, leaving a watch unplaced at episode termination (\texttt{t = 1,059}); (c) a suboptimal post-interruption recovery choice in \emph{Put Bottles Into Dustbin} (L2), where the planner set down a still-grasped yellow bottle outside the bin and incurred a costly regrasp; and (d) an unresolved slip alert in \emph{Build Tower} (L2), where the robot spent its final 38 steps rotating in Fine-step mode without placing the upper board.

Overall, runtime monitoring reliably returns control early upon predicted risk, missing perception, or realized disturbance, while ultimate task recovery further depends on tracking stability, predicate coverage, physical regrasp capability, and timely high-level recovery planning.
\end{document}